\documentclass{article}

\usepackage{microtype}
\usepackage{graphicx}
\usepackage{booktabs}
\usepackage{multirow}
\usepackage{placeins}
\usepackage{amsmath}
\usepackage{amssymb}
\usepackage{xcolor}
\usepackage{tikz}
\usepackage{pgfplots}
\pgfplotsset{compat=1.18}
\usetikzlibrary{arrows.meta,patterns,positioning,fit,calc,decorations.pathreplacing}
\usepackage{algorithm}
\usepackage{algorithmic}
\usepackage{hyperref}

\usepackage{mlsys2025}
\renewcommand{\icmlruler}[1]{}

\newcommand{\papertitle}{Scaling VLA Serving Under Robotic Execution Constraints}
\newcommand{\papertitleshort}{XInfer}
\newcommand{\sys}{\textsc{XInfer}}

\newcommand{\modelA}{Wall-OSS-0.5}
\newcommand{\modelB}{$\pi_{0.5}$}

\newif\ifdraft
\drafttrue

\newcommand{\Nstar}{N^{*}}
\newcommand{\TE}{T_{E}}
\newcommand{\ms}{\ensuremath{\,\mathrm{ms}}}
\newcommand{\stageName}[1]{\text{\rmfamily\upshape #1}}
\newcommand{\phaseV}{\stageName{ViT}}
\newcommand{\phaseP}{\stageName{Prefill}}
\newcommand{\phaseA}{\stageName{Action}}

\mlsystitlerunning{\papertitleshort}

\begin{document}

\twocolumn[
\mlsystitle{\papertitle}

\begin{center}
    {\bfseries X Square Robot Team}
\end{center}

\mlsysaffiliation{anon}{Anonymous Institution}
\mlsyscorrespondingauthor{Anonymous Authors}{anon@example.com}

\mlsyskeywords{Machine Learning Systems, Model Serving, Robot Learning, Vision-Language-Action Models, GPU Sharing}

\vskip 0.3in

\begin{abstract}
Vision-language-action (VLA) models have emerged as a promising paradigm for general-purpose robotic manipulation. In our production setting, hundreds of VLA-powered robots are already operating in real-world environments, creating an immediate need for efficient fleet serving. We present \sys{}, a production-deployed VLA serving system designed to reduce the inference infrastructure cost of large-scale robotic deployments.
\sys{} jointly coordinates heterogeneous VLA computation with the evolving response requirements of robotic execution. By exploiting computational parallelism while managing interference, \sys{} improves GPU efficiency while protecting time-critical robotic feedback.
We evaluate \sys{} across three VLA models using production-derived workloads, achieving \textbf{1.26–3.75×} the effective robot capacity (ERC) of state-of-the-art systems under physical-execution constraints. Our implementation is publicly available at
\url{https://github.com/X-Square-Robot/xinfer}.
\end{abstract}
\begin{minipage}{\textwidth}
\centering
\includegraphics[width=\textwidth]{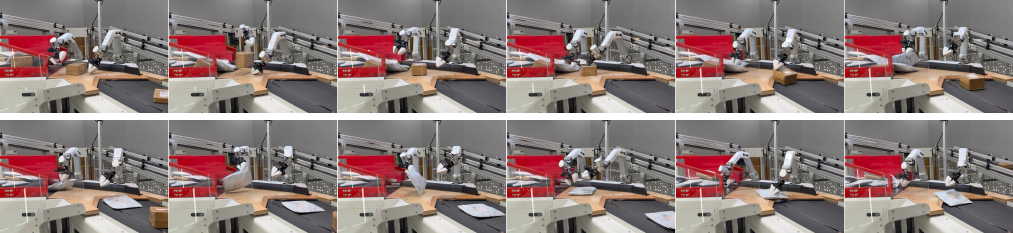}
\par\vspace{-10pt}
\makeatletter
\def\@captype{figure}
\makeatother
\caption{Production parcel handling. A carton (top) and a soft parcel (bottom)
are reoriented and pushed onto the conveyor. Each row follows the same parcel
from left to right.}
\label{fig:production-sequences}
\end{minipage}

\vspace{6pt}
]


\section{Introduction}
\label{sec:intro}

Vision-language-action (VLA) models are bringing general-purpose robotic manipulation into logistics and manufacturing, where multiple robots can share GPU infrastructure to reduce serving cost. Figure~\ref{fig:production-sequences} shows representative parcel-handling operations from our production deployment. Recent inference engines improve the efficiency of individual VLA queries \citep{flashrt,embodiedcpp,flashvla}. However, higher inference throughput does not necessarily translate into greater physical productivity, as delayed feedback can disrupt robotic execution. We use \emph{effective robot capacity} (ERC), the maximum number of concurrent robots a GPU can serve while meeting specified latency and reliability requirements, as the serving objective.

\begin{figure}[!t]
\centering
\input{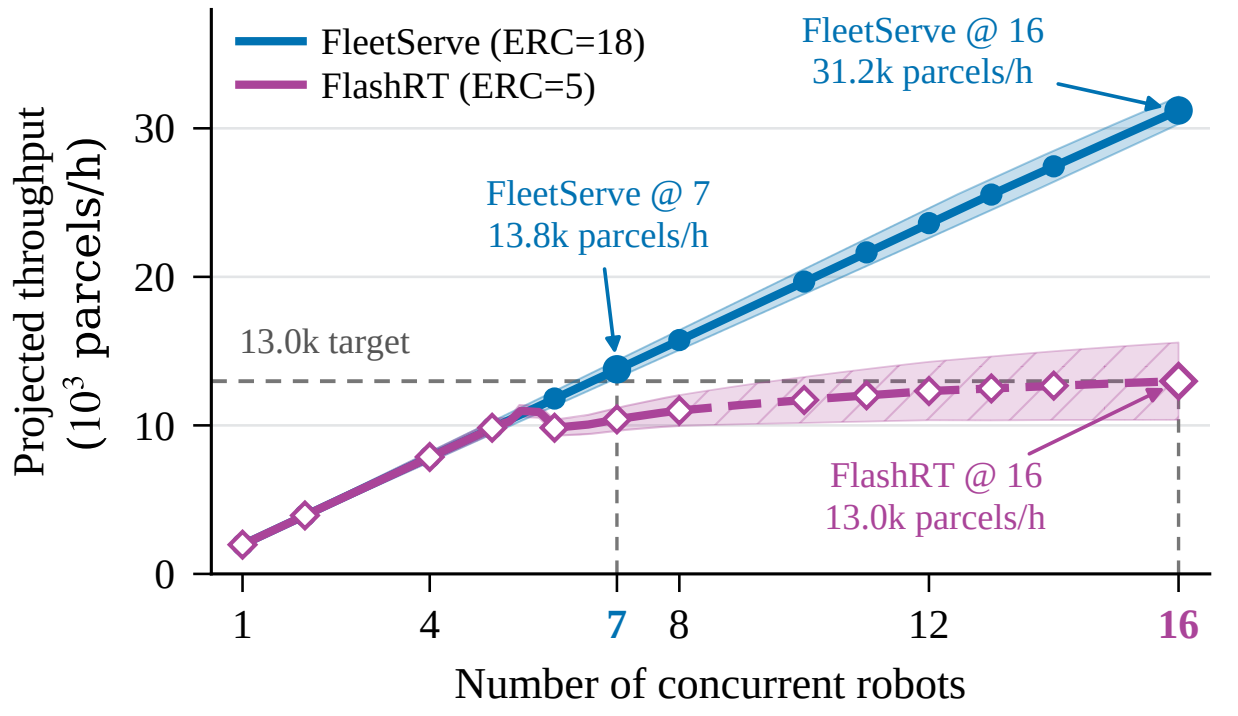}
\par\vspace{-10pt}
\caption{Projected parcel throughput on one GPU (VLA-2B), combining measured
latency with single-robot trials. Guides mark 13k parcels/h; ERC requires response
P50 $<150\ms{}$ and zero request errors. Shading reflects trial variation; dashed curves and hatching
mark extrapolation.}
\label{fig:teaser}
\end{figure}

Shared VLA serving requires exploiting concurrency without compromising time-critical feedback. VLA requests traverse heterogeneous stages with distinct computational demands, creating opportunities to overlap independent work across robots. However, concurrent GPU execution also introduces interference: admitting a new stage may delay stages already in progress. Since these stages are non-preemptive, prioritizing urgent requests alone cannot prevent deadline violations caused by previously admitted work. Effective fleet serving therefore requires two coordinated decisions: which requests should be prioritized, and which stages can be admitted without jeopardizing the deadlines of current and forthcoming requests.

The cost of this interference is not uniform across requests. Robots continue executing previously predicted actions while awaiting new responses, and the consequences of delayed feedback depend on their ongoing physical execution. As a result, requests with similar computational demands may have substantially different response-time requirements, which evolve throughout task execution. Moreover, response timing influences subsequent request arrivals, coupling scheduling decisions with future workload dynamics. Fleet serving must therefore balance concurrency against interference under changing per-request deadlines, rather than optimize inference throughput or average latency alone.

Prior systems adapt action horizons and request priorities to execution context \citep{kairos,armory}, coordinate serving for factory productivity \citep{rosa}, or overlap encoding and diffusion in lockstep rounds \citep{robion}. Yet effective fleet serving must reconcile the time-varying urgency of robotic feedback with the distinct computational demands of VLA stages. This calls for fine-grained scheduling that matches stage execution to available GPU capacity while controlling interference with time-critical requests.

We present \sys{}, a VLA serving system that couples asynchronous execution with fine-grained, execution-aware scheduling. Its dependency-driven engine exposes parallelism across requests and heterogeneous VLA stages, allowing computation to progress as dependencies resolve. The scheduler coordinates these execution opportunities with the time-varying urgency of robotic feedback by decoupling request prioritization from stage admission. At stage boundaries, it combines per-request response-time budgets, remaining computation, and predicted interference to admit ordinary work only when protected requests remain predicted feasible. It further accounts for anticipated earlier-deadline arrivals using request history, balancing immediate concurrency gains against future scheduling demands.

We evaluate VLA-2B, $\pi_{0.5}$, and Wall-OSS-0.5
using production-derived closed-loop workloads and controlled experiments. With error-free
responses, \sys{} achieves ERCs of 18 VLA-2B and
24 $\pi_{0.5}$ robots, versus 10 and 15 for vLLM-Omni;
Robion serves 19 $\pi_{0.5}$ robots.
In a separate VLA-2B evaluation with heterogeneous response-time budgets, all policies share the same backend and execution resources. Under the same real-time feasibility criterion, \sys{} achieves an ERC of 18 robots, compared with 15 for Slack-Gate, 14 for FIFO, and 12 for EDF (\S\ref{sec:eval}). Mechanism ablations isolate the effects of asynchronous execution and scheduling decisions.

Our contributions are:
\begin{itemize}
\setlength\itemsep{1pt}
\setlength\topsep{2pt}
\item \textbf{An asynchronous VLA serving engine} that exposes cross-request
and cross-stage parallelism by advancing work as its dependencies complete.
\item \textbf{Execution-aware cross-stage scheduling} that coordinates shared
GPU computation by separating request priority from stage admission, accounting
for interference on current and anticipated urgent requests.
\item \textbf{Production deployment and empirical evaluation.} \sys{} is deployed
in production. We characterize closed-loop workloads, compare ERC across
three VLA backends, and evaluate heterogeneous-budget scheduling and its mechanisms.
\end{itemize}

\section{Background}
\label{sec:workload}

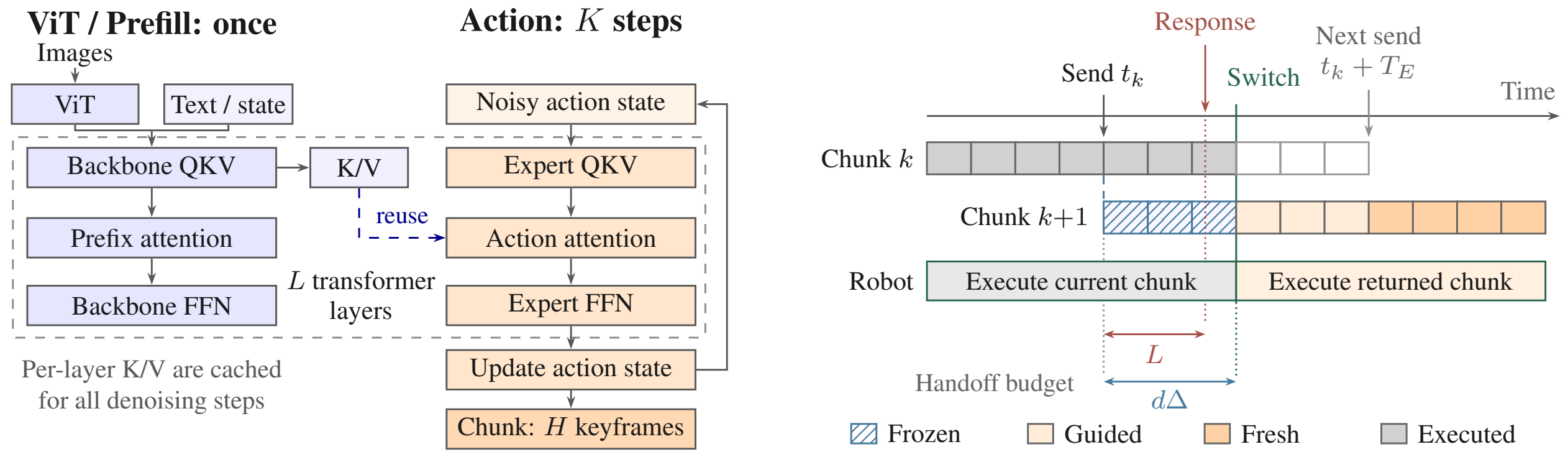
\begin{figure*}[t]
\centering
\begin{minipage}[b]{0.485\textwidth}
\centering
\begin{tikzpicture}[x=.96cm,y=-.90cm,font=\fontsize{8}{9}\selectfont,
  text=black!90,line width=.55pt,inner sep=1pt,
  flow/.style={-{Stealth[length=1.3mm,width=.9mm]},draw=black!65},
  block/.style={draw=black!65,minimum height=.42cm,align=center,inner sep=2pt},
  prefixblk/.style={block,fill=blue!10,minimum width=2.65cm},
  expert/.style={block,fill=orange!20,minimum width=2.65cm}]
\path[use as bounding box] (0,0) rectangle (8.25,5.25);
\node[font=\bfseries] at (1.70,.20) {ViT / Prefill: once};
\node[font=\bfseries] at (6.35,.20) {Action: $K$ steps};
\node at (.85,.58) {Images};
\node[block,fill=blue!10,minimum width=1.35cm] (vit) at (.85,1.16) {ViT};
\node[block,fill=blue!5,minimum width=1.35cm] (txt) at (2.55,1.16) {Text / state};
\node[expert,fill=orange!10] (noisy) at (6.35,1.16) {Noisy action state};
\draw[flow] (.85,.73)--(vit.north);
\draw[black!65] (vit.south)--(.85,1.47)--(1.70,1.47);
\draw[black!65] (txt.south)--(2.55,1.47)--(1.70,1.47);

\node[prefixblk] (pqkv) at (1.70,1.91) {Backbone QKV};
\node[expert] (aqkv) at (6.35,1.91) {Expert QKV};
\node[prefixblk] (patt) at (1.70,2.74) {Prefix attention};
\node[expert] (aatt) at (6.35,2.74) {Action attention};
\node[prefixblk] (pffn) at (1.70,3.54) {Backbone FFN};
\node[expert] (affn) at (6.35,3.54) {Expert FFN};
\draw[black!45,dashed,rounded corners=1pt]
  (.16,1.55) rectangle (7.84,3.92);
\node[align=center,inner sep=0pt,font=\fontsize{8}{9}\selectfont]
  at (4.02,3.45) {$L$ transformer\\layers};
\draw[flow] (1.70,1.47)--(pqkv.north);
\draw[flow] (noisy.south)--(aqkv.north);
\draw[flow] (pqkv.south)--(patt.north);
\draw[flow] (patt.south)--(pffn.north);
\draw[flow] (aqkv.south)--(aatt.north);
\draw[flow] (aatt.south)--(affn.north);

\node[block,fill=blue!6,minimum width=1.04cm] (kv) at (4.00,1.91) {K/V};
\draw[flow] (pqkv.east)--(kv.west);
\draw[flow,draw=blue!55!black,dashed] (kv.south)--(4.00,2.74)--(aatt.west);
\node[text=blue!55!black,font=\fontsize{7.5}{8.5}\selectfont] at (4.48,2.50) {reuse};
\node[align=center,text=black!65,font=\fontsize{7.8}{9.4}\selectfont]
  at (1.70,4.50) {Per-layer K/V are cached\\for all denoising steps};

\node[expert] (update) at (6.35,4.30) {Update action state};
\node[expert,fill=orange!30] (chunk) at (6.35,5.00) {Chunk: $H$ keyframes};
\draw[flow] (affn.south)--(update.north);
\draw[flow] (update.south)--(chunk.north);
\draw[flow] (update.east)--(8.08,4.30)--(8.08,1.16)--(noisy.east);
\end{tikzpicture}\unskip\par\vspace{1mm}
{\small (a) VLA computation and KV reuse}
\end{minipage}\hfill
\begin{minipage}[b]{0.485\textwidth}
\centering
\begin{tikzpicture}[x=.96cm,y=-.90cm,font=\fontsize{8}{9}\selectfont,
  text=black!90,line width=.55pt,inner sep=1pt,
  flow/.style={-{Stealth[length=1.3mm,width=.9mm]},draw=black!65}]
\path[use as bounding box] (0,0) rectangle (8.25,5.25);
\def\u{.49}
\def\xzero{1.10}
\def\send{3.06}
\def\response{4.187}
\def\handoff{4.53}
\def\nextsend{6.00}
\definecolor{rtcBlue}{HTML}{477A9F}
\definecolor{rtcGreen}{HTML}{276651}
\definecolor{rtcRed}{HTML}{A45348}

\draw[flow] (\xzero,1.30)--(8.12,1.30);
\node[anchor=south east,text=black!60] at (8.12,1.16) {Time};
\node[anchor=south] at (\send,1.00) {Send $t_k$};
\draw[flow] (\send,1.05)--(\send,1.55);
\node[anchor=south,text=rtcRed] at (\response,.40) {Response};
\draw[flow,draw=rtcRed] (\response,.46)--(\response,1.26);
\node[anchor=south west,text=rtcGreen,xshift=-2mm] at (\handoff+.06,1.00) {Switch};
\draw[rtcGreen] (\handoff,1.16)--(\handoff,3.50);
\node[anchor=south,align=center,text=black!60] at (\nextsend,.91)
  {Next send\\$t_k+\TE$};
\draw[flow,draw=black!45] (\nextsend,.99)--(\nextsend,1.55);

\node[anchor=east] at (1.00,1.80) {Chunk $k$};
\foreach \i in {0,...,9}{
  \ifnum\i<7
    \filldraw[fill=black!18,draw=black!60]
      (\xzero+\i*\u,1.61) rectangle ({\xzero+(\i+1)*\u},1.99);
  \else
    \draw[black!40] (\xzero+\i*\u,1.61) rectangle ({\xzero+(\i+1)*\u},1.99);
  \fi
}
\node[anchor=east] at (\send-.12,2.50) {Chunk $k{+}1$};
\foreach \i in {0,...,9}{
  \ifnum\i<3
    \path[fill=blue!3] (\send+\i*\u,2.31) rectangle ({\send+(\i+1)*\u},2.69);
    \path[pattern=north east lines,pattern color=rtcBlue]
      (\send+\i*\u,2.31) rectangle ({\send+(\i+1)*\u},2.69);
    \draw[rtcBlue] (\send+\i*\u,2.31) rectangle ({\send+(\i+1)*\u},2.69);
  \else
    \ifnum\i<6
      \filldraw[fill=orange!15,draw=black!60]
        (\send+\i*\u,2.31) rectangle ({\send+(\i+1)*\u},2.69);
    \else
      \filldraw[fill=orange!35,draw=black!60]
        (\send+\i*\u,2.31) rectangle ({\send+(\i+1)*\u},2.69);
    \fi
  \fi
}
\draw[rtcBlue,densely dashed] (\send,1.99)--(\send,2.28);
\draw[rtcRed,densely dotted] (\response,1.34)--(\response,3.75);
\draw[black!45,densely dotted] (\send,2.74)--(\send,4.37);
\draw[rtcGreen,densely dotted] (\handoff,3.50)--(\handoff,4.37);

\node[anchor=east] at (1.00,3.25) {Robot};
\filldraw[fill=black!9,draw=rtcGreen]
  (\xzero,3.02) rectangle (\handoff,3.48);
\filldraw[fill=orange!12,draw=rtcGreen]
  (\handoff,3.02) rectangle (7.96,3.48);
\node[font=\fontsize{7.8}{9}\selectfont] at (2.815,3.25) {Execute current chunk};
\node[font=\fontsize{7.8}{9}\selectfont] at (6.245,3.25) {Execute returned chunk};

\draw[{Stealth[length=1.15mm]}-{Stealth[length=1.15mm]},rtcRed]
  (\send,3.88)--(\response,3.88);
\node[anchor=north,text=rtcRed] at (3.6235,3.96) {$L$};
\draw[{Stealth[length=1.15mm]}-{Stealth[length=1.15mm]},rtcBlue]
  (\send,4.44)--(\handoff,4.44);
\node[anchor=north,text=rtcBlue] at (3.795,4.49) {$d\Delta$};
\node[anchor=east,text=black!60,font=\fontsize{7.7}{8.7}\selectfont] at (2.78,4.48)
  {Handoff budget};

\path[pattern=north east lines,pattern color=rtcBlue] (.26,4.94) rectangle (.54,5.17);
\draw[rtcBlue] (.26,4.94) rectangle (.54,5.17);
\node[anchor=west] at (.62,5.055) {Frozen};
\filldraw[fill=orange!15,draw=black!60] (2.22,4.94) rectangle (2.50,5.17);
\node[anchor=west] at (2.58,5.055) {Guided};
\filldraw[fill=orange!35,draw=black!60] (4.18,4.94) rectangle (4.46,5.17);
\node[anchor=west] at (4.54,5.055) {Fresh};
\filldraw[fill=black!18,draw=black!60] (6.14,4.94) rectangle (6.42,5.17);
\node[anchor=west] at (6.50,5.055) {Executed};
\end{tikzpicture}\unskip\par\vspace{1mm}
{\small (b) Real-time chunking}
\end{minipage}
\caption{VLA computation and real-time chunking. (a) Conditioning keys and values are
reused across denoising steps. (b) The frozen prefix aligns the next chunk with
committed actions while inference overlaps execution. The illustrated early response
satisfies $L<d\Delta<\TE$, so the next send occurs at $t_k+\TE$.}
\label{fig:bg}
\end{figure*}

\subsection{Heterogeneous VLA Computation}
\label{sec:bg:vla}

Vision-language-action (VLA) models map observations and instructions to robot
actions. We study models coupling a vision-language backbone to an action expert
that generates continuous actions through flow matching or diffusion
\citep{flowmatching,diffusionpolicy,pi05,walloss05,walloss}. As Figure~\ref{fig:bg}(a)
shows, the backbone computes conditioning keys and values once; the expert reuses
them over $K$ denoising steps to generate an \emph{action chunk} of $H$ future
keyframes \citep{act}. We distinguish three computational stages: ViT encodes
visual observations, Prefill performs language-model computation after visual
encoding, and Action generates actions through denoising. These stages differ
in computation and memory-access patterns. The expert's weights differ from
the backbone's, and its evolving action state is private to each request.

\subsection{Closed-Loop Robot Control}
\label{sec:workload:period}

Real-time chunking (RTC) \citep{rtc} aligns independently generated chunks
by freezing the next chunk's first $d$ keyframes to committed actions
(Figure~\ref{fig:bg}(b)). For a request issued at $t_k$ with keyframe
spacing $\Delta$, the planned handoff is $t_k+d\Delta$ and its budget is
$d\Delta$. A late chunk can cause an action discontinuity if applied without
alignment to current execution progress. If the action queue empties, our
controller pauses execution until a replacement chunk becomes usable.

Longer commitment permits more inference time but postpones feedback, even
when $d$ adapts to recent latency (\S\ref{sec:motivation:existing}).
Appendix~\ref{app:serving:rtc} gives the inpainting and handoff details.

\subsection{Fleet Workload and Request Contract}
\label{sec:workload:fleet}
\label{sec:workload:contract}

A fleet shares inference resources across robots, each running its own control loop.
A request carries observations, robot state, and model inputs. Each request $j$
is associated with a response-time budget $b_j$ that expresses its feedback
requirement. The budget is supplied as request metadata and may be assigned by
the robot controller, a task planner, or a policy at the serving frontend.
RTC requests additionally supply the aligned previous action chunk and
commitment length $d$. The response budget, RTC handoff budget
$d\Delta$, and minimum request period $\TE$ are distinct quantities: they specify
the serving requirement, the planned chunk handoff, and the earliest next send.

Robots continue their current chunks while requests are processed. We use a
minimum-period request policy with at most one outstanding request per robot.
Let $s_j$ be the actual send time of request $j$ and $r_j$ its response arrival time.
Abstracting local client-side handling, the next
request is sent at
\begin{equation}
s_{j+1} = \max(s_j + \TE, r_j).
\label{eq:request-trigger}
\end{equation}
The next send occurs when the minimum interval has elapsed and the previous response
has arrived. Thus, $N$ robots offer at most $N/\TE$ requests per unit time;
late responses reduce both offered load and action-update frequency.

\begin{figure}[t]
\centering
\begingroup
\definecolor{tracebusy}{HTML}{0072B2}
\definecolor{tracesend}{HTML}{555B65}
\definecolor{traceload}{HTML}{27867B}
\pgfplotsset{fleettrace/.style={
  scale only axis, width=0.78\columnwidth, axis lines=left,
  axis line style={black!65, line width=0.45pt, -},
  tick align=outside, tick style={black!65, line width=0.4pt},
  tick label style={font=\fontsize{7.5}{8.5}\selectfont},
  label style={font=\fontsize{8.5}{9.5}\selectfont},
  xmin=0, xmax=2000, xtick={0,500,...,2000}, xmajorgrids=true,
  grid style={black!10, line width=0.35pt}, clip=false}}
\begin{tikzpicture}
\begin{axis}[fleettrace, name=streams, height=0.11\columnwidth, unbounded coords=jump,
  ymin=0.5, ymax=5.5, ytick={1,5}, yticklabels={S1,S5}, y dir=reverse,
  xticklabels={}]
\addplot[color=tracebusy, line width=1.9pt, line cap=butt]
  table[x=ms, y=stream, col sep=comma] {figures/data/fleet-zoom.csv};
\addplot[only marks, mark=|, mark size=1.7pt, color=tracesend, mark options={line width=0.5pt}]
  table[x=ms, y=stream, col sep=comma] {figures/data/fleet-zoom-sends.csv};
\end{axis}
\begin{axis}[fleettrace, at={(streams.south west)}, anchor=north west, yshift=-3pt,
  height=0.19\columnwidth, ymin=0, ymax=5.4, ytick={0,1,...,5}, ymajorgrids=true,
  xlabel={Time (ms)}, xlabel style={yshift=2pt}, ylabel={In flight}, ylabel style={yshift=-3pt}]
\addplot[const plot, draw=traceload, line width=0.8pt, fill=traceload!18]
  table[x=ms, y=inflight, col sep=comma] {figures/data/fleet-inflight.csv} \closedcycle;
\draw[densely dashed, black!70, line width=0.6pt] (axis cs:0,2.45) -- (axis cs:2000,2.45)
  node[anchor=west, font=\fontsize{7.5}{8.5}\selectfont, align=left, inner sep=1.5pt] {mean\\2.45};
\end{axis}
\end{tikzpicture}
\endgroup
\caption{Five separately recorded request streams aligned on a common 2-s axis.
Top: bars mark requests in flight and ticks mark sends.
Bottom: the superimposed in-flight count.}
\label{fig:fleet-trace}
\end{figure}
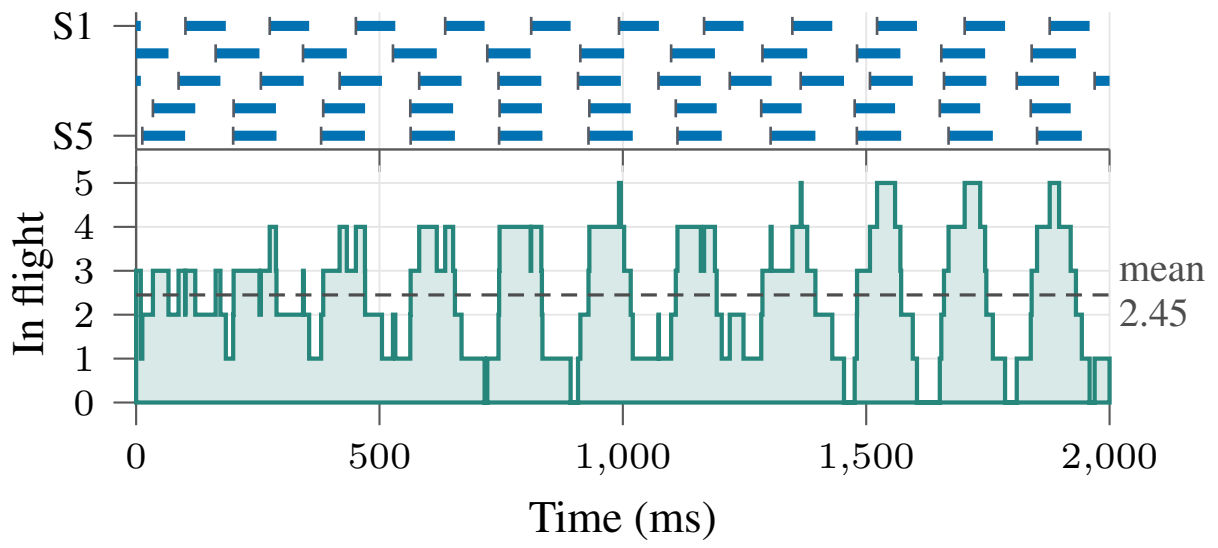

Figure~\ref{fig:fleet-trace} overlays five streams from deployed robots,
including two sessions from one robot. Requests occupy only 45--53\% of each
dedicated instance's timeline; combined arrivals show gaps and bursts.

\paragraph{Objective and service metrics.}
We use \emph{effective robot capacity} (ERC), denoted $\Nstar$, to measure the
maximum number of concurrent robots a GPU can serve while meeting specified
latency and reliability requirements. Each comparison fixes the model, workload,
and acceptance criterion. Response latency measures feedback delay.
SLO attainment is the fraction of responses meeting their supplied budgets;
\emph{on-time goodput} counts such responses per unit time.
Section~\ref{sec:eval:setup} specifies the acceptance criteria and measurement
windows for each experiment.

\section{Motivation}
\label{sec:motivation}
\label{sec:space}
\label{sec:bg:motivation}

Sharing GPU capacity across independent robots must preserve timely feedback.
The workload in Section~\ref{sec:workload} motivates three requirements: express
changing feedback needs, expose independent work, and control concurrent interference.

\subsection{Feedback Needs Change During Execution}
\label{sec:motivation:existing}
\label{sec:bg:existing}

In our delay-injection study, adding 200\ms{} before action delivery reduces
completed parcel output by 29.6\%. Separate fixed-count studies observe additional
attempts for parcel reorientation and conveyor pushing, both contact operations.
These results connect delayed feedback to reduced physical progress;
Appendix~\ref{app:physical-latency} gives the protocols and results.

Acceptable delay varies during execution. Contact operations, such as grasping,
parcel flipping, and conveyor pushing, or interaction with a changing scene may
require frequent updates. Predictable free-space transport may tolerate a longer
execution horizon. Increasing the RTC commitment length buys response time but
postpones feedback; it cannot replace timely service when rapid adaptation is
needed. A fixed priority per robot cannot capture these changes.

These observations call for per-request feedback requirements. Planned motion,
execution progress, and application-specific signals, such as gripper events or
contact observations, provide cues about how quickly feedback is needed.
A budget-selection policy can map this execution context to a response-time
budget for each request. The contract in Section~\ref{sec:workload:contract}
exposes these budgets to the scheduler through a uniform interface.

\paragraph{Closed-loop throughput and goodput.}
By Equation~\ref{eq:request-trigger}, robot $i$'s $k$th response latency
$L_{i,k}$ determines send interval $\tau_{i,k}=\max(\TE,L_{i,k})$.
Over robot $i$'s requests, define mean cycle overrun $h_i=\mathbb E[(L_i-\TE)^+]$
and attainment $p_i=\Pr(L_i\le b_i)$, where $b_i$ is the request's
budget and $(x)^+=\max(x,0)$. For $N$ continuously active robots with stationary,
ergodic request sequences and finite mean latency, long-run throughput $X$
and on-time goodput $J$ satisfy
\begin{equation}
X=\sum_{i=1}^{N}\frac{1}{\TE+h_i},
\qquad
J=\sum_{i=1}^{N}\frac{p_i}{\TE+h_i}.
\label{eq:closed-loop-rates}
\end{equation}
These long-run rates follow from one response per mean cycle $\TE+h_i$,
with fraction $p_i$ on time.
Delaying wide-budget requests can lengthen request cycles even when responses
remain on time. Higher attainment can therefore coexist with lower goodput;
scheduling must preserve update frequency as well as deadline attainment.
Section~\ref{sec:eval:slo} evaluates both effects.

\subsection{Independent Requests Offer Parallelism}
\label{sec:motivation:opportunities}

The streams in Figure~\ref{fig:fleet-trace} leave gaps in each dedicated instance
yet produce bursts when combined. Sharing can fill these gaps by coordinating
independent work. Each request follows ViT, Prefill, and Action dependencies,
but CPU preparation can overlap GPU execution, and another request's ViT or
Prefill can progress during Action. Serializing complete requests or advancing stages in common rounds
introduces waits that these dependencies do not require.

At 24 robots, enabling the full execution engine on \modelB{} raises
throughput from 54.90 to 131.53\,req/s (2.40$\times$,
Figure~\ref{fig:serving:cumulative}). These results motivate exposing work
as its dependencies complete.

\subsection{Throughput Alone Does Not Protect Deadlines}
\label{sec:motivation:challenges}

Throughput gains need not improve tail latency. In a separate 32-robot
\modelB{} ablation, adding independent progression to multi-stream execution
raises QPS by 9.7\%, but P99 rises from 467.07 to 560.28\ms{}
(Table~\ref{tab:serving:mechanism-details}).

A wide-budget request's Prefill can consume GPU resources needed by a tight-budget
Action. Deadline ordering \citep{liulayland} chooses which queued request runs
next, but moving the urgent request to the front does not release resources held
by non-preemptive work. Urgency must account for remaining computation and
interference, and starting another stage must depend on its effect on that work.

Protecting only requests already present can also be too late: a long-running
stage may still be active when an earlier-deadline request arrives. In matched
experiments with alternating task phases, arrival-aware admission raises mean
SLO attainment by 5.0 and 11.3 percentage points at 16 and 18 robots,
respectively (Table~\ref{tab:slo:gpu-ablation}). Conversely,
holding all other work whenever a request becomes urgent can forgo useful overlap.

The scheduler must therefore separate request protection from stage admission.
Admission should permit useful overlap while keeping the deadlines of current
and anticipated urgent requests predicted feasible. Urgency alone should not stop
all other work.


\section{Design}
\label{sec:design}

\sys{} separates the availability of parallel work from the decision to execute it.
An asynchronous engine makes work ready as its dependencies complete.
A central scheduler identifies requests that need protection, then admits
ordinary work only when the resulting interference preserves predicted
deadline feasibility. This division addresses the tension in
Section~\ref{sec:motivation}: exploiting parallelism without delaying urgent
action updates.

\subsection{System Overview}
\label{sec:design:overview}

The server consumes per-request response-time budgets and RTC inputs through
the request contract in Section~\ref{sec:workload:contract}. Budget assignment
and execution-horizon selection are separate from the scheduler's request
prioritization and stage admission.

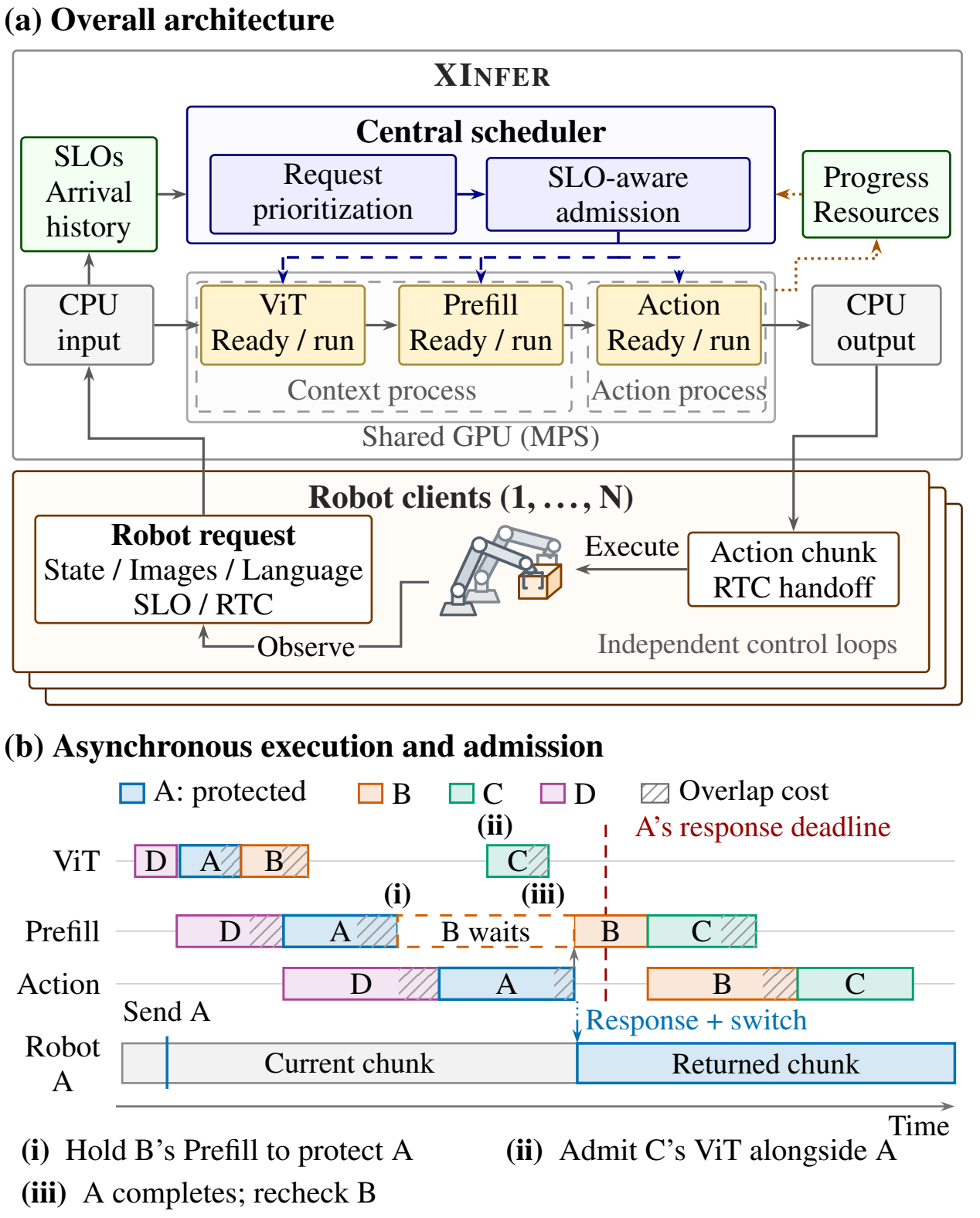
\begin{figure}[t]
\centering
\begin{minipage}{\columnwidth}
\fontsize{8}{9}\selectfont
\raggedright\textbf{(a) Overall architecture}\par\vspace{1.5pt}
\centering
\definecolor{gpuStageFill}{HTML}{FFF4CF}
\definecolor{gpuStageEdge}{HTML}{A88D3D}
\begin{tikzpicture}[x=1cm,y=1cm,
 font=\fontsize{7.5}{8.5}\selectfont,
 box/.style={draw=black!50,line width=.5pt,rounded corners=1pt,
   align=center,inner sep=2pt},
 data/.style={-{Stealth[length=1.3mm,width=.9mm]},draw=black!65,line width=.6pt},
 control/.style={-{Stealth[length=1.3mm,width=.9mm]},draw=blue!55!black,
   line width=.65pt,dashed},
 feedback/.style={-{Stealth[length=1.3mm,width=.9mm]},draw=orange!65!black,
   line width=.65pt,densely dotted}]
\path[use as bounding box] (0,-.62) rectangle (8.20,5.04);

\draw[box,fill=white,draw=black!45] (.04,1.50) rectangle (8.16,5.00);
\node[font=\fontsize{8}{9}\selectfont\bfseries,text=black!85] at (4.14,4.79) {\sys{}};

\begin{scope}[yshift=6.2mm]
\draw[box,fill=blue!2,draw=blue!45!black] (1.53,2.73) rectangle (6.56,3.90);
\node[font=\fontsize{8}{9}\selectfont\bfseries] at (4.045,3.69) {Central scheduler};
\node[box,fill=blue!7,draw=blue!45!black,minimum width=2.10cm,minimum height=.58cm]
 (priority) at (2.78,3.15) {Request\\prioritization};
\node[box,fill=blue!7,draw=blue!45!black,minimum width=2.25cm,minimum height=.58cm]
 (admission) at (5.22,3.15) {SLO-aware\\admission};
\draw[data,draw=blue!55!black] (priority.east)--(admission.west);
\node[box,fill=green!5,draw=green!30!black,minimum width=1.16cm,minimum height=.74cm]
 (demand) at (.69,3.15) {SLOs\\Arrival\\history};
\node[box,fill=green!5,draw=green!30!black,minimum width=1.26cm,minimum height=.74cm]
 (state) at (7.43,3.15) {Progress\\Resources};
\draw[data] (demand.east)--(1.53,3.15);
\draw[feedback] (state.west)--(6.56,3.15);
\end{scope}

\begin{scope}[yshift=10mm]
\draw[box,fill=black!1,draw=black!35] (1.53,.82) rectangle (6.56,2.10);
\draw[box,dashed,draw=black!40] (1.61,.91) rectangle (4.82,2.02);
\draw[box,dashed,draw=black!40] (4.96,.91) rectangle (6.48,2.02);
\node[font=\fontsize{7.2}{8.2}\selectfont,text=black!65] at (3.20,1.07) {Context process};
\node[font=\fontsize{7.2}{8.2}\selectfont,text=black!65] at (5.74,1.07) {Action process};
\node[font=\fontsize{7.3}{8.3}\selectfont,text=black!70] at (4.045,.68) {Shared GPU (MPS)};
\node[box,fill=black!4,minimum width=1.10cm,minimum height=.59cm] (cpuin)
 at (.69,1.65) {CPU\\input};
\node[box,fill=gpuStageFill,draw=gpuStageEdge,minimum width=1.40cm,minimum height=.52cm]
 (vit) at (2.35,1.65) {ViT\\Ready / run};
\node[box,fill=gpuStageFill,draw=gpuStageEdge,minimum width=1.40cm,minimum height=.52cm]
 (prefill) at (4.045,1.65) {Prefill\\Ready / run};
\node[box,fill=gpuStageFill,draw=gpuStageEdge,minimum width=1.40cm,minimum height=.52cm]
 (action) at (5.74,1.65) {Action\\Ready / run};
\node[box,fill=black!4,minimum width=1.10cm,minimum height=.59cm] (cpuout)
 at (7.43,1.65) {CPU\\output};
\draw[data] (cpuin.east)--(vit.west);
\draw[data] (vit.east)--(prefill.west);
\draw[data] (prefill.east)--(action.west);
\draw[data] (action.east)--(cpuout.west);
\end{scope}

\draw[draw=blue!55!black,line width=.65pt,dashed]
 (admission.south)--(5.22,3.23)--(2.35,3.23);
\draw[control] (2.35,3.23)--(vit.north);
\draw[control] (4.045,3.23)--(prefill.north);
\draw[control] (5.22,3.23)--(5.74,3.23)--(action.north);
\draw[feedback] (6.56,2.95)--(6.74,2.95)--(6.74,3.22)--(7.43,3.22)--(state.south);

\draw[box,fill=orange!1,draw=orange!30!black] (.32,-.60) rectangle (8.16,1.12);
\draw[box,fill=orange!2,draw=orange!35!black] (.18,-.46) rectangle (8.02,1.26);
\draw[box,fill=orange!3,draw=orange!40!black] (.04,-.32) rectangle (7.88,1.40);
\node[font=\fontsize{8}{9}\selectfont\bfseries,text=black!85] at (3.96,1.16) {Robot clients (1, \ldots, N)};
\begin{scope}[yshift=-1.8mm]
\node[box,fill=white,draw=orange!45!black,minimum width=2.70cm,minimum height=.80cm]
 (request) at (1.67,.74) {\textbf{Robot request}\\State / Images / Language\\SLO / RTC};
\node[box,fill=white,draw=orange!45!black,minimum width=1.80cm,minimum height=.63cm]
 (response) at (6.72,.74) {Action chunk\\RTC handoff};
\begin{scope}[shift={(3.57,.31)}]
\definecolor{robotEdge}{HTML}{53616D}
\definecolor{robotLink}{HTML}{CED6DC}
\begin{scope}[line join=round,line cap=round]
  \path (0,0) rectangle (1.25,.85);
  \draw[draw=robotEdge!75!white,fill=black!10,line width=.65pt] (0.330000,0.150000)--(0.630000,0.150000)--(0.590000,0.250000)--(0.370000,0.250000)--cycle;
  \draw[draw=robotEdge!75!white,fill=black!7,line width=.65pt] (0.430000,0.240000) rectangle (0.530000,0.370000);
  \draw[draw=robotEdge!75!white,fill=robotLink,line width=.72pt] (0.433780,0.384899)--(0.637711,0.763464)--(0.730151,0.713667)--(0.526220,0.335101)--cycle;
  \draw[draw=robotEdge!75!white,fill=robotLink!65!white,line width=.72pt] (0.697186,0.789365)--(1.113255,0.680799)--(1.086745,0.579201)--(0.670676,0.687767)--cycle;
  \draw[draw=robotEdge!75!white,fill=robotLink!75!white,line width=.72pt] (1.128000,0.609000)--(1.068000,0.529000)--(1.012000,0.571000)--(1.072000,0.651000)--cycle;
  \draw[draw=robotEdge!75!white,fill=white,line width=.65pt] (0.480000,0.360000) circle (0.057);
  \draw[draw=robotEdge!75!white,fill=white,line width=.65pt] (0.683931,0.738566) circle (0.06);
  \draw[draw=robotEdge!75!white,fill=white,line width=.65pt] (1.100000,0.630000) circle (0.051);
  \draw[draw=robotEdge!75!white,line width=.72pt] (0.960000,0.550000)--(1.120000,0.550000)
    (0.960000,0.550000)--(0.960000,0.420000)--(0.985000,0.420000)
    (1.120000,0.550000)--(1.120000,0.420000)--(1.095000,0.420000);
  \draw[draw=orange!50!black,fill=orange!8,line width=.55pt]
    (.80,.37)--(1.04,.37)--(1.16,.47)--(.92,.47)--cycle;
  \draw[draw=orange!50!black,fill=orange!28,line width=.55pt]
    (1.04,.18)--(1.16,.28)--(1.16,.47)--(1.04,.37)--cycle;
  \draw[draw=orange!50!black,fill=orange!15,line width=.55pt]
    (.80,.18) rectangle (1.04,.37);
  \draw[draw=robotEdge!100!white,fill=black!10,line width=.65pt] (0.130000,0.040000)--(0.430000,0.040000)--(0.390000,0.140000)--(0.170000,0.140000)--cycle;
  \draw[draw=robotEdge!100!white,fill=black!7,line width=.65pt] (0.230000,0.130000) rectangle (0.330000,0.260000);
  \draw[draw=robotEdge!100!white,fill=robotLink,line width=.72pt] (0.232373,0.272088)--(0.413285,0.662179)--(0.508540,0.618003)--(0.327627,0.227912)--cycle;
  \draw[draw=robotEdge!100!white,fill=robotLink!65!white,line width=.72pt] (0.480458,0.688816)--(0.879546,0.528726)--(0.840454,0.431274)--(0.441366,0.591365)--cycle;
  \draw[draw=robotEdge!100!white,fill=robotLink!75!white,line width=.72pt] (0.888000,0.501000)--(0.948000,0.421000)--(0.892000,0.379000)--(0.832000,0.459000)--cycle;
  \draw[draw=robotEdge!100!white,fill=white,line width=.65pt] (0.280000,0.250000) circle (0.057);
  \draw[draw=robotEdge!100!white,fill=white,line width=.65pt] (0.460912,0.640091) circle (0.06);
  \draw[draw=robotEdge!100!white,fill=white,line width=.65pt] (0.860000,0.480000) circle (0.051);
  \draw[draw=robotEdge!100!white,line width=.72pt] (0.840000,0.400000)--(1.000000,0.400000)
    (0.840000,0.400000)--(0.840000,0.270000)--(0.865000,0.270000)
    (1.000000,0.400000)--(1.000000,0.270000)--(0.975000,0.270000);
\end{scope}
\end{scope}
\draw[data] (response.west)--(4.85,.74);
\node[font=\fontsize{7.2}{8.2}\selectfont,fill=orange!3,inner sep=.5pt]
 at (5.34,.94) {Execute};
\draw[data] (3.60,.63)--(3.36,.63)--(3.36,.08)--(1.67,.08)--(request.south);
\node[font=\fontsize{7.2}{8.2}\selectfont,fill=orange!3,inner sep=.7pt]
 at (2.55,.08) {Observe};
\node[font=\fontsize{7}{8}\selectfont,text=black!65,anchor=east,inner sep=0pt]
 at (7.62,.08) {Independent control loops};
\end{scope}
\draw[data] (request.north)--(1.67,1.68)--(.69,1.68)--(cpuin.south);
\draw[data] (cpuout.south)--(7.43,1.70)--(6.72,1.70)--(response.north);
\draw[data] (cpuin.north)--(demand.south);
\end{tikzpicture}\par\vspace{3pt}
\raggedright\textbf{(b) Asynchronous execution and admission}\par\vspace{1.5pt}
\centering
\begin{tikzpicture}[x=1cm,y=1cm,
 font=\fontsize{7.5}{8.5}\selectfont,
 edge/.style={-{Stealth[length=1.2mm,width=.8mm]},draw=black!55,line width=.5pt},
 ra/.style={draw=reqA,fill=reqA!16,line width=.65pt},
 rb/.style={draw=reqB,fill=reqB!15,line width=.55pt},
 rc/.style={draw=reqC,fill=reqC!15,line width=.55pt},
 rd/.style={draw=reqD,fill=reqD!14,line width=.55pt},
 decision/.style={font=\fontsize{7.2}{8.2}\selectfont\bfseries,inner sep=.3pt}]
\definecolor{reqA}{HTML}{0072B2}
\definecolor{reqB}{HTML}{D55E00}
\definecolor{reqC}{HTML}{009E73}
\definecolor{reqD}{HTML}{AA4499}
\path[use as bounding box] (0,0) rectangle (8.2,3.95);
\draw[ra] (.96,3.70) rectangle (1.16,3.86);
\node[anchor=west,inner sep=1pt] at (1.20,3.78) {A: protected};
\draw[rb] (3.00,3.70) rectangle (3.20,3.86);
\node[anchor=west,inner sep=1pt] at (3.24,3.78) {B};
\draw[rc] (3.78,3.70) rectangle (3.98,3.86);
\node[anchor=west,inner sep=1pt] at (4.02,3.78) {C};
\draw[rd] (4.56,3.70) rectangle (4.76,3.86);
\node[anchor=west,inner sep=1pt] at (4.80,3.78) {D};
\draw[draw=black!60,pattern=north east lines,pattern color=black!40]
 (5.42,3.70) rectangle (5.65,3.86);
\node[anchor=west,inner sep=1pt,font=\fontsize{7.2}{8.2}\selectfont]
 at (5.70,3.78) {Overlap cost};
\draw[red!60!black,densely dashed,line width=.55pt]
 (5.1128,2.00) -- (5.1128,3.53);
\node[anchor=west,inner sep=1pt,text=red!60!black,
 font=\fontsize{7.2}{8.2}\selectfont]
 at (5.32,3.47) {A's response deadline};
\draw[black!18,line width=.35pt] (.92,3.20) -- (8.1,3.20);
\node[anchor=east,inner sep=1pt] at (.83,3.20) {ViT};
\draw[black!18,line width=.35pt] (.92,2.60) -- (8.1,2.60);
\node[anchor=east,inner sep=1pt] at (.83,2.60) {Prefill};
\draw[black!18,line width=.35pt] (.92,2.15) -- (8.1,2.15);
\node[anchor=east,inner sep=1pt] at (.83,2.15) {Action};
\draw[rd] (1.0846,3.0650) rectangle (1.4409,3.3350);
\node[inner sep=.2pt] at (1.2627,3.20) {D};
\draw[rd] (1.4409,2.4650) rectangle (2.3548,2.7350);
\draw[draw=none,pattern=north east lines,pattern color=black!38] (2.0707,2.4650) rectangle (2.3548,2.7350);
\node[inner sep=.2pt] at (1.8978,2.60) {D};
\draw[rd] (2.3548,2.0150) rectangle (3.6873,2.2850);
\draw[draw=none,pattern=north east lines,pattern color=black!38] (3.3414,2.0150) rectangle (3.6873,2.2850);
\node[inner sep=.2pt] at (3.0210,2.15) {D};
\draw[ra] (1.4681,3.0650) rectangle (1.9921,3.3350);
\draw[draw=none,pattern=north east lines,pattern color=black!38] (1.8243,3.0650) rectangle (1.9921,3.3350);
\node[inner sep=.2pt] at (1.7301,3.20) {A};
\draw[ra] (2.3548,2.4650) rectangle (3.3310,2.7350);
\draw[draw=none,pattern=north east lines,pattern color=black!38] (2.9852,2.4650) rectangle (3.3310,2.7350);
\node[inner sep=.2pt] at (2.8429,2.60) {A};
\draw[ra] (3.6873,2.0150) rectangle (4.8417,2.2850);
\draw[draw=none,pattern=north east lines,pattern color=black!38] (4.6739,2.0150) rectangle (4.8417,2.2850);
\node[inner sep=.2pt] at (4.2645,2.15) {A};
\draw[rb] (1.9921,3.0650) rectangle (2.5656,3.3350);
\draw[draw=none,pattern=north east lines,pattern color=black!38] (2.3483,3.0650) rectangle (2.5656,3.3350);
\node[inner sep=.2pt] at (2.2788,3.20) {B};
\draw[rb] (4.8417,2.4650) rectangle (5.4721,2.7350);
\node[inner sep=.2pt] at (5.1569,2.60) {B};
\draw[rb] (5.4721,2.0150) rectangle (6.7551,2.2850);
\draw[draw=none,pattern=north east lines,pattern color=black!38] (6.4587,2.0150) rectangle (6.7551,2.2850);
\node[inner sep=.2pt] at (6.1136,2.15) {B};
\draw[rc] (4.0990,3.0650) rectangle (4.6230,3.3350);
\draw[draw=none,pattern=north east lines,pattern color=black!38] (4.4552,3.0650) rectangle (4.6230,3.3350);
\node[inner sep=.2pt] at (4.3610,3.20) {C};
\draw[rc] (5.4721,2.4650) rectangle (6.3988,2.7350);
\draw[draw=none,pattern=north east lines,pattern color=black!38] (6.1025,2.4650) rectangle (6.3988,2.7350);
\node[inner sep=.2pt] at (5.9355,2.60) {C};
\draw[rc] (6.7551,2.0150) rectangle (7.7418,2.2850);
\node[inner sep=.2pt] at (7.2484,2.15) {C};
\draw[draw=reqB,dashed,fill=white,line width=.55pt]
 (3.3310,2.465) rectangle (4.8417,2.735);
\node[font=\fontsize{7.2}{8.2}\selectfont,inner sep=0pt]
 at (4.0864,2.60) {B waits};
\node[decision] at (3.3310,2.90) {(i)};
\node[decision] at (4.1663,3.49) {(ii)};
\node[decision,anchor=east] at (4.7848,2.90) {(iii)};
\draw[edge] (4.8417,2.285) -- (4.8417,2.465);
\node[anchor=east,align=center,inner sep=1pt] at (.83,1.47) {Robot\\A};
\draw[draw=black!45,fill=black!5,line width=.5pt]
 (0.9747,1.30) rectangle (4.8689,1.64);
\node[font=\fontsize{7.2}{8.2}\selectfont] at (2.9218,1.47)
 {Current chunk};
\draw[ra] (4.8689,1.30) rectangle (8.10,1.64);
\node[font=\fontsize{7.2}{8.2}\selectfont] at (6.4844,1.47)
 {Returned chunk};
\draw[reqA,line width=.65pt] (1.3582,1.25) -- (1.3582,1.70);
\node[anchor=south,font=\fontsize{7.2}{8.2}\selectfont]
 at (1.3582,1.73) {Send A};
\draw[reqA,densely dotted,line width=.5pt]
 (4.8689,2.00) -- (4.8689,1.84);
\draw[edge,draw=reqA] (4.8689,1.84) -- (4.8689,1.64);
\node[anchor=south west,inner sep=1pt,text=reqA,
 font=\fontsize{7.2}{8.2}\selectfont]
 at (4.8989,1.66) {Response + switch};
\draw[edge] (.92,1.10) -- (8.10,1.10);
\node[anchor=north east,inner sep=1pt,font=\fontsize{7.2}{8.2}\selectfont]
 at (8.1,1.07) {Time};
\node[anchor=west,inner sep=0pt,font=\fontsize{7.5}{8.5}\selectfont] at (.10,.70)
 {\textbf{(i)}\enspace Hold B's Prefill to protect A};
\node[anchor=west,inner sep=0pt,font=\fontsize{7.5}{8.5}\selectfont] at (4.25,.70)
 {\textbf{(ii)}\enspace Admit C's ViT alongside A};
\node[anchor=west,inner sep=0pt,font=\fontsize{7.5}{8.5}\selectfont] at (.10,.34)
 {\textbf{(iii)}\enspace A completes; recheck B};
\end{tikzpicture}\par
\end{minipage}
\caption{\sys{} architecture and illustrative execution.
(a) Central scheduling controls admission at the ViT, Prefill, and Action entries.
(b) Admission permits concurrent work when A's response is predicted to meet its deadline.}
\label{fig:system-overview}
\end{figure}

Figure~\ref{fig:system-overview}(a) shows the ViT, Prefill, and Action pipeline,
preceded by CPU preparation and followed by output processing. The engine
maintains dependencies, ready queues, and bounded execution resources;
the central scheduler controls admission at each stage entry. Solid arrows carry data,
dashed arrows convey admission decisions, and dotted arrows return scheduler state.

The policy makes two decisions: \emph{which requests need protection}
(\S\ref{sec:design:priority}) and \emph{which ready stages may start}
(\S\ref{sec:design:slo}). Both use request progress and shared resource state;
the latter also considers arrival history. Issued GPU units are non-preemptive.
Section~\ref{sec:impl} specifies each backend's execution boundaries.

In Figure~\ref{fig:system-overview}(b), A has an urgent Action, B a ready Prefill,
and C additional vision work; D is preceding ordinary work.

\subsection{Asynchronous Stage Execution}
\label{sec:design:execution}
\label{sec:design:prefill}
\label{sec:design:handoff}

\paragraph{Advance work by its dependencies.}
Each request follows ViT, Prefill, and Action dependencies without a common
round barrier. Once Prefill supplies its conditioning state, Action can run
alongside another request's ViT or Prefill. CPU preparation and output
processing can overlap GPU work from other requests. Admission controls
interference before additional GPU work starts.

\paragraph{Separate submission paths.}
ViT and \phaseP{} share a process; \phaseA{} runs in another, with independent
host submission loops and private execution state. This avoids serializing their
host work through one submission loop. CUDA streams support concurrency within
the ViT/Prefill process; MPS \citep{mps} permits the processes to execute
concurrently on the shared GPU. Requests reuse read-only weights and retain
private intermediate state. Bounded state slots propagate backpressure when a
downstream stage cannot accept work.

\paragraph{Dispatch admitted work.}
\label{sec:design:action}
The engine dispatches admitted work at backend-supported execution boundaries.
Each execution unit advances on its own completion events, without a common
round barrier, and runs to completion. Request state remains private.
Section~\ref{sec:impl} maps computational stages to execution units and
specifies their dispatch boundaries.

\begin{figure*}[!t]
\centering
\begin{tikzpicture}[x=1cm,y=-.85cm,
  font=\fontsize{8}{9.2}\selectfont,text=black!90,
  line width=.5pt,inner sep=1pt,
  arrow/.style={-{Stealth[length=1.25mm,width=.85mm]},draw=black!65},
  panel/.style={font=\fontsize{8.6}{10}\selectfont\bfseries},
  secondary/.style={text=black!65,font=\fontsize{7.7}{8.7}\selectfont}]
\definecolor{schedBlue}{HTML}{477A9F}
\definecolor{schedPale}{HTML}{E9F0F5}
\definecolor{schedGold}{HTML}{F4D477}
\definecolor{schedGoldEdge}{HTML}{A88D3D}
\definecolor{schedRed}{HTML}{A45348}
\definecolor{schedLightRed}{HTML}{F5E7E3}
\definecolor{schedGray}{HTML}{D5DCE1}
\path[use as bounding box] (0,0) rectangle (17.14,4.73);
\draw[black!16,line width=.35pt] (5.61,.10)--(5.61,4.24);
\draw[black!16,line width=.35pt] (11.47,.10)--(11.47,4.24);

\begin{scope}
\node at (2.70,.25) {A: $\ell_A-t=50$ ms, $R_A=40$ ms};
\node[anchor=east] at (1.01,.83) {Slack};
\filldraw[fill=schedGold,draw=schedGoldEdge]
  (1.15,.63) rectangle (3.35,1.03);
\node at (2.25,.83) {$\sigma_A=10$};
\node[secondary] at (2.70,1.42) {Ordinary running units (remaining ms)};
\draw[schedGoldEdge,densely dashed,line width=.65pt]
  (3.35,.56)--(3.35,1.13);
\draw[schedGoldEdge,densely dashed,line width=.65pt]
  (3.35,1.72)--(3.35,3.30);
\node[anchor=east] at (1.01,1.95) {ViT};
\node[anchor=east] at (1.01,2.48) {Prefill};
\node[anchor=east] at (1.01,3.01) {Action};
\foreach \cy/\endx/\value in {1.95/2.03/4,2.48/4.23/14,3.01/2.91/8}{
  \filldraw[fill=schedGray,draw=black!60]
    (1.15,\cy-.20) rectangle (\endx,\cy+.20);
  \node[anchor=west] at (\endx+.08,\cy) {\value};
}
\draw[black!80,line width=.85pt] (1.15,2.28) rectangle (4.23,2.68);
\node at (2.70,3.60) {$B_A=\max(4,14,8)=14$ ms};
\node[font=\fontsize{8.4}{9.4}\selectfont] at (2.70,4.05)
  {$\sigma_A<B_A\ \Rightarrow$ \textbf{Protect A}};
\node[panel] at (2.70,4.52) {(a) Priority promotion};
\end{scope}

\begin{scope}[xshift=5.86cm]
\node at (2.70,.25) {Protected A: shared slack = 10 ms};
\node[anchor=west] at (.24,.82) {1. Check Prefill $u$};
\filldraw[fill=schedBlue,draw=schedBlue]
  (.40,1.15) rectangle (2.98,1.60);
\filldraw[fill=schedGold,draw=schedGoldEdge]
  (2.98,1.15) rectangle (4.70,1.60);
\node[text=white] at (1.69,1.375) {6 extra delay};
\node at (3.84,1.375) {4 left};
\node[anchor=west] at (.40,1.98) {$\Delta_A(u)=6\leq10$};
\node[anchor=east,text=schedBlue,font=\bfseries] at (4.70,1.98) {Admit $u$};
\draw[arrow] (2.55,2.22)--(2.55,2.73);
\node[anchor=west,secondary] at (2.72,2.47) {Refresh state};
\node[anchor=west] at (.24,2.94) {2. Check ViT $v$};
\filldraw[fill=schedGold,draw=schedGoldEdge]
  (.40,3.27) rectangle (2.12,3.72);
\fill[schedLightRed] (2.12,3.27) rectangle (2.98,3.72);
\path[pattern=north east lines,pattern color=schedRed!65]
  (2.12,3.27) rectangle (2.98,3.72);
\draw[schedRed,line width=.7pt] (.40,3.27) rectangle (2.98,3.72);
\draw[schedGoldEdge,densely dashed] (2.12,3.12)--(2.12,3.87);
\node at (1.26,3.495) {4 left};
\node[fill=schedLightRed,text=schedRed,inner sep=.3pt] at (2.55,3.495) {$+2$};
\node[anchor=west] at (.40,4.05) {$\Delta_A(v)=6>4$};
\node[anchor=east,text=schedRed,font=\bfseries] at (4.70,4.05) {Defer $v$};
\node[panel] at (2.70,4.52) {(b) Shared-slack admission};
\end{scope}

\begin{scope}[xshift=11.72cm]
\node at (2.70,.25) {$w$ passes the current-request check};
\node[secondary] at (2.70,.72) {Forecast $e$ has an earlier deadline: $\ell_e<\ell_w$};
\draw[arrow] (.65,1.32)--(5.17,1.32);
\node[anchor=south] at (.65,1.28) {$t$};
\node[anchor=south,text=schedRed] at (3.90,1.28) {$\ell_e$};
\node[anchor=south] at (5.00,1.28) {$\ell_w$};
\draw[schedRed,densely dashed,line width=.65pt] (3.90,1.43)--(3.90,3.01);
\draw[black!45,densely dashed] (5.00,1.43)--(5.00,3.01);
\node[anchor=east] at (.50,1.88) {$w$};
\filldraw[fill=schedGray,draw=black!60] (.65,1.68) rectangle (3.15,2.08);
\node at (1.90,1.88) {If admitted now};
\node[anchor=east] at (.50,2.66) {$e$};
\fill[schedPale] (1.45,2.43) rectangle (4.65,2.89);
\fill[schedLightRed] (3.30,2.43) rectangle (4.65,2.89);
\path[pattern=north east lines,pattern color=schedRed!65]
  (3.30,2.43) rectangle (4.65,2.89);
\draw[schedBlue,dashed,line width=.65pt]
  (1.45,2.43) rectangle (4.65,2.89);
\node[text=schedBlue] at (2.35,2.66) {Forecast};
\draw[schedBlue,densely dotted] (1.45,2.20)--(1.45,3.12);
\node[anchor=north,secondary] at (1.45,3.20) {Arrival};
\draw[black!55] (3.30,3.02)--(3.30,3.10)--(4.65,3.10)--(4.65,3.02);
\node[anchor=north,secondary,align=center] at (3.98,3.20) {Extra delay};
\node[font=\fontsize{8.4}{9.4}\selectfont] at (2.70,4.05)
  {Predicted miss $\Rightarrow$ {\color{schedRed}\textbf{Defer $w$}}};
\node[panel] at (2.70,4.52) {(c) Arrival-aware admission};
\end{scope}
\end{tikzpicture}\unskip\par\vspace{-4pt}
\caption{Scheduling decisions (illustrative predictions).
(a) Protect A using remaining slack. (b) Refresh shared slack after each launch.
(c) Defer $w$ for an earlier-deadline forecast.
Panel (b) fixes $t$ and $H=\{A\}$, with additive costs and other checks passing.
Dashed outlines mark forecasts; hatching marks excess cost in (b) and extra delay in (c).}
\label{fig:scheduling-decisions}
\end{figure*}
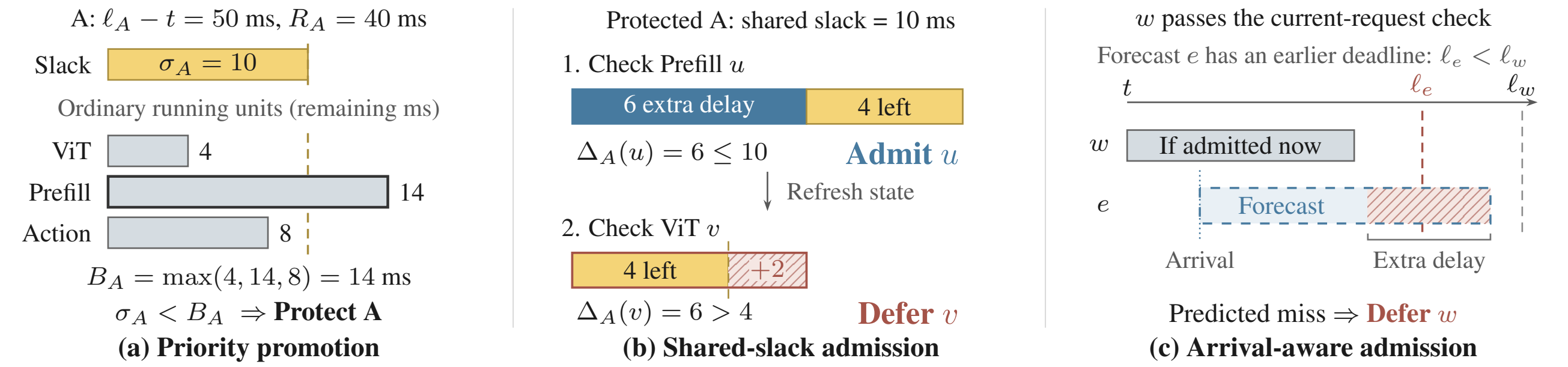

\subsection{Request Prioritization}
\label{sec:design:priority}

The scheduler protects queued and running requests according to their remaining
slack and the work that could delay them.

\paragraph{Progress-aware service-time estimation.}
At server arrival $a_j$, request $j$ with budget $b_j$ receives scheduling
deadline $\ell_j=a_j+b_j-\delta$, where $\delta$ reserves estimated time
outside the server. Given projected concurrent stage counts $\mathbf n$,
a calibrated model $R_j(\mathbf n)$ estimates remaining service time, covering
the current stage's residual, subsequent stages, and output processing,
excluding future queueing.
At time $t$, the slack available for waiting and additional interference is
\begin{equation}
\sigma_j(\mathbf n)=\ell_j-t-R_j(\mathbf n).
\label{eq:slo-slack}
\end{equation}
The predictor tracks each running stage's progress by accumulating calibrated
execution rates as concurrency changes, and refreshes $R_j$ at each decision.
Appendix~\ref{app:slo:model} specifies the model and calibration;
Appendix~\ref{app:slo:online-prediction} evaluates remaining-time prediction
for running stages.

\paragraph{Blocking-aware promotion.}
Following non-preemptive response-time analysis \citep{rta,nonpreemptive}, let
$B_j$ be the maximum predicted remaining time of another request's running
ordinary unit, or zero if none exists. When $\sigma_j<B_j$, waiting for that
unit could exhaust $j$'s slack, so the scheduler promotes $j$.
In Figure~\ref{fig:scheduling-decisions}(a), A is promoted because its
10\ms{} slack is below $B_A=14$\ms{}.
Promotion persists while the request is live. Effective protection is withheld
while the request is predicted late even in isolation; it remains eligible for
service and can regain protection if the prediction changes.

\textsc{RefreshPriority} returns the effective protected set $H$, including
queued and running requests, and foreseeable load $\mathbf n_f$: running units
plus protected ready work that available resources can issue. Ready requests
are ordered by protected status and then deadline.

\subsection{SLO-Aware Admission}
\label{sec:design:deadline}
\label{sec:design:slo}

Protected work starts when resources permit; ordinary candidates must pass
current-request and forecast checks to limit interference on urgent work.
This adapts slack stealing
\citep{slackstealing} to concurrent GPU stages, using parallel capacity when
the affected deadlines remain predicted feasible.

\paragraph{Current-request protection.}
Consider an ordinary candidate unit $u$ at stage $s$, not yet included in
foreseeable load $\mathbf n_f$. At a fixed decision time, let
$\mathbf n^+=\mathbf n_f+\mathbf e_s$ add that unit. Its predicted interference
on protected request $h$ and the resulting slack satisfy
\begin{equation}
\begin{split}
\Delta_h(u)&=R_h(\mathbf n^+)-R_h(\mathbf n_f),\\
\sigma_h(\mathbf n^+)&=\sigma_h(\mathbf n_f)-\Delta_h(u).
\end{split}
\label{eq:admission-interference-budget}
\end{equation}
\textsc{CurrentFeasible} accepts exactly when
$\Delta_h(u)\leq\sigma_h(\mathbf n_f)$ for every $h\in H$.

\emph{Local property.} A passing candidate leaves the current protected set
predicted feasible under this decision's snapshot.
\emph{Proof.} The inequality gives $\sigma_h(\mathbf n^+)\geq0$, hence
$t+R_h(\mathbf n^+)\leq\ell_h$ for every $h\in H$.

A failed check leaves the candidate queued while other ready work is considered.

\paragraph{Coordinate decisions across stages.}
Arrivals, completions, and periodic ticks trigger Algorithm~\ref{alg:slo-policy}.
The driver visits \phaseA{}, \phaseP{}, and ViT. \textsc{ScanReady} starts from
priority-ordered ready work and yields candidates while the stage has capacity,
retaining deferred work for later scans. Each launch updates shared execution state and refreshes
protection before the next check.
Figure~\ref{fig:scheduling-decisions}(b) shows why: under the illustrative
additive predictor, admitting $u$ consumes 6\ms{} of A's 10\ms{} slack,
leaving 4\ms{}. Candidate $v$ would add another 6\ms{} and is deferred.
Independent checks against the original snapshot would admit both.
Appendix~\ref{app:slo:model} specifies the priority update and stage
iteration rules.

\paragraph{Arrival-aware admission.}
In Figure~\ref{fig:scheduling-decisions}(c), $w$ passes the current-request
check but would delay forecast request $e$ past its earlier deadline.
\textsc{ForecastFeasible} uses recent arrivals grouped by response budget
to forecast requests during the candidate's execution, retaining those whose
deadlines precede the candidate's. Each retained forecast must satisfy
$\delta+f+R_V(\mathbf n'_e)\leq b_e$, where $f$ reserves front-end time,
$R_V$ estimates the remaining pipeline from ViT, and $\mathbf n'_e$ includes
the candidate and preceding retained forecasts.

\begin{algorithm}[t]
\caption{SLO-aware stage dispatch}
\label{alg:slo-policy}
\small
\begin{algorithmic}[1]
\REQUIRE State $S$: ready queues, running work, arrival history
\STATE Record arrivals and completions in $S$
\FOR{$s\in[\phaseA{},\phaseP{},\phaseV{}]$}
  \STATE $(H,\mathbf n_f)\gets$ \textsc{RefreshPriority}$(S)$
  \FOR{$u$ in \textsc{ScanReady}$(s,S,H)$}
    \IF{$\operatorname{request}(u)\notin H$}
      \STATE $\mathbf n^+\gets\mathbf n_f+\mathbf e_s$
      \IF{\textbf{not} \textsc{CurrentFeasible}$(H,\mathbf n^+)$}
        \STATE \textbf{continue} \COMMENT{protect current requests}
      \ENDIF
      \IF{\textbf{not} \textsc{ForecastFeasible}$(u,\mathbf n^+)$}
        \STATE \textbf{continue} \COMMENT{protect forecast arrivals}
      \ENDIF
    \ENDIF
    \STATE \textsc{Issue}$(u)$; update queues and running state in $S$
    \STATE $(H,\mathbf n_f)\gets$ \textsc{RefreshPriority}$(S)$
  \ENDFOR
\ENDFOR
\end{algorithmic}
\end{algorithm}

\section{Implementation}
\label{sec:impl}

Model adapters connect input preparation, kernels, CUDA graphs, and output
processing to the runtime. VLA-2B and \modelA{} use MXFP8 ViT/Prefill linear
operations \citep{microscaling} and BF16 Action; \modelB{} uses native FlashRT
FP8 with explicit state-text conditioning (Appendix~\ref{app:serving:protocol}).
The central scheduler dispatches each stage separately; the three-model
backends combine ViT and Prefill as Context. Each Action dispatch runs a
complete, non-preemptive denoising trajectory with fixed request membership;
central Action dispatches contain one request.
Client timestamps measure response latency; server timers and event traces
diagnose execution. Runtime controls toggle GPU-stage overlap, cohort barriers,
and CPU--GPU overlap (Section~\ref{sec:eval:ablation}).
Appendix~\ref{app:slo:model} records the central predictor's calibration and constants.

\section{Evaluation}
\label{sec:eval}

\subsection{Experimental Setup}
\label{sec:eval:setup}

\paragraph{Models and serving comparisons.}
Each serving run uses NVIDIA consumer-grade Blackwell GPUs. The three-model study covers VLA-2B,
\modelA{}, and \modelB{}. VLA-2B uses a Qwen3.5-2B backbone
and non-public weights. We use requests representative of our deployment setting
and keep inputs fixed across serving methods within each model. Each request
contains three camera views, a shared language instruction, and a
14-dimensional synthetic robot state. Table~\ref{tab:serving:inputs} summarizes
the input shapes and actual prefix lengths. All models encode 868 image patches
into 217 visual tokens; text lengths reflect their native tokenizers and templates.
Appendix~\ref{app:serving:protocol} details input preprocessing.
For each model, we compare \sys{} with integrations of SGLang \citep{sglang}
and vLLM-Omni \citep{vllmomni}, FlashRT (served through its Nexus frontend) \citep{flashrt},
and a Robion single-GPU reimplementation \citep{robion}. The latter combines
true stage batches, lockstep progression, and calibrated Context SM quotas;
Context group $g$ overlaps Action group $g-1$, with a barrier between rounds.
Comparisons share the checkpoint, workload, precision, and compute backend.
Appendix~\ref{app:serving:robion} lists the Robion reproduction settings;
Appendix~\ref{app:serving:protocol} provides the measurement protocol.

\begin{table}[t]
\centering\footnotesize
\setlength{\tabcolsep}{3.5pt}
\caption{Input configuration. Token segments sum to the full prefix;
action tokens and sequence padding are excluded.}
\label{tab:serving:inputs}
\begin{tabular}{@{}lrrr@{}}
\toprule
Input & VLA-2B & Wall-OSS-0.5 & $\pi_{0.5}$ \\
\midrule
Camera 1 (pixels) & $224\!\times\!352$ & $196\!\times\!308$ & $196\!\times\!308$ \\
Cameras 2/3 (pixels) & $224\!\times\!320$ & $196\!\times\!280$ & $196\!\times\!280$ \\
Patch size (pixels) & $16\!\times\!16$ & $14\!\times\!14$ & $14\!\times\!14$ \\
\addlinespace[3pt]
Visual tokens & 217 & 217 & 217 \\
Instruction tokens & 41 & 41 & 40 \\
State tokens & 72 & 74 & 127 \\
Template/special tokens & 116 & 113 & 11 \\
Full prefix & \textbf{446} & \textbf{445} & \textbf{395} \\
\bottomrule
\end{tabular}
\end{table}

\begin{figure*}[t]
\centering
\includegraphics[width=\textwidth]{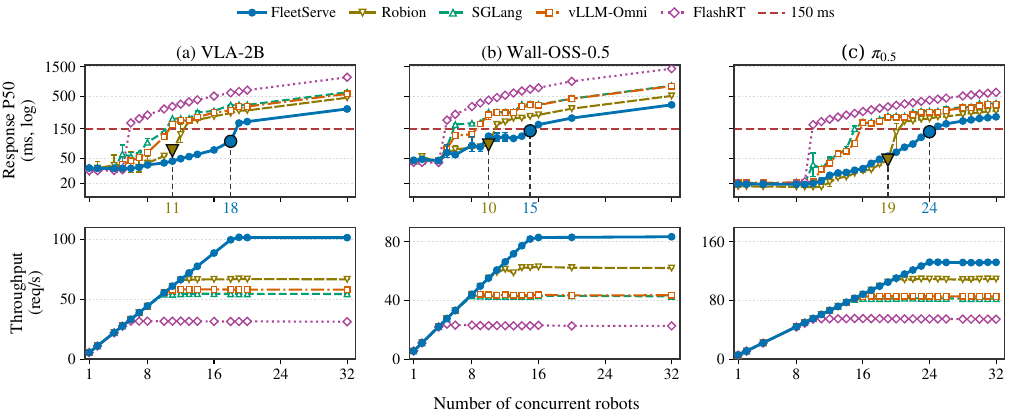}
\caption{Serving performance across three models. Latency uses a log scale; dashed lines mark
150\ms{}. Large markers and guides identify confirmed ERCs for \sys{} and Robion.}
\label{fig:serving-capacity}
\end{figure*}

\paragraph{Central-scheduling comparison.}
All SLO policies share the three-stage VLA-2B backend, model, kernels, and
allocated GPU resources. FIFO orders ready requests by arrival. EDF orders each
stage queue by deadline with fixed per-stage concurrency limits. Slack-Gate
uses deadline queues and defers non-urgent work when another request has less
than 40\ms{} estimated slack. Appendix~\ref{app:slo:final} details baseline
configurations and parameter selection. Policy-specific bounds and admission
decisions determine actual concurrency. \sys{} applies the same prediction
and admission rules across workloads. Matched policy comparisons use the same GPU.

\paragraph{Load and metrics.}
Each emulated robot follows a closed-loop request stream with at most one request
in flight and a minimum send interval of 180\ms{} (Equation~\ref{eq:request-trigger}). Response latency spans sending a
packed request to receiving the complete response; capture, packing, and parsing
are excluded. The three-model runs use ten denoising steps and 16 future action
points. The RTC commitment length is fixed at $d=8$ in both studies; response-time
budgets in the central-scheduling replay follow each workload.
Both serving studies measure requests over 180\,s after
30\,s of warmup. We report ERC as the largest validated integer robot population.
The three-model comparison requires zero request errors and response P50 below
150\ms{} in all three windows; the central-scheduling study requires at least 95\% SLO attainment.
Three-model QPS counts successful requests sent and completed within
the window; central-scheduling goodput counts on-time responses to requests sent
within it. Both rates divide these counts by 180\,s.
Appendix~\ref{app:serving:protocol} details aggregation and measurement-window
boundaries; Appendices~\ref{app:serving:capacity} and~\ref{app:slo:final}
detail ERC validation.

\paragraph{ERC criteria and physical task throughput.}
We apply the deployment-derived 150\ms{} threshold across models:
50\ms{} baseline response plus 100\ms{} injected delay, near the
delay--throughput knee (Appendix~\ref{app:physical-budget}). For heterogeneous
budgets, the 95\% attainment target defines a common high-attainment operating
point allowing at most 5\% deadline misses; goodput captures the accompanying
response-rate tradeoff. Figure~\ref{fig:teaser} projects fleet parcel throughput
from measured serving latency and physical calibration
(Appendix~\ref{app:physical-projection}).

\begin{table}[t]
\centering
\small
\caption{ERC (robots) under the response-latency criterion.}
\label{tab:serving-capacity}
\setlength{\tabcolsep}{0.5pt}
\newlength{\capacityMethodWidth}
\newlength{\capacityModelWidth}
\settowidth{\capacityMethodWidth}{Serving method}
\settowidth{\capacityModelWidth}{FlashRT}
\ifdim\capacityModelWidth>\capacityMethodWidth
\setlength{\capacityMethodWidth}{\capacityModelWidth}
\fi
\setlength{\capacityModelWidth}{\dimexpr(\columnwidth-\capacityMethodWidth-6\tabcolsep)/3\relax}
\newcommand{\capacityModelHeader}[1]{\parbox[c]{\capacityModelWidth}{\centering #1}}
\begin{tabular}{@{}lccc@{}}
\toprule
Serving method & \capacityModelHeader{VLA-2B} & \capacityModelHeader{\mbox{Wall-OSS-0.5}} & \capacityModelHeader{$\pi_{0.5}$} \\
\midrule
\sys{} & \textbf{18} & \textbf{15} & \textbf{24} \\
SGLang & 10 & 5 & 14 \\
vLLM-Omni & 10 & 8 & 15 \\
FlashRT & 5 & 4 & 9 \\
Robion & 11 & 10 & 19 \\
\bottomrule
\end{tabular}
\end{table}

\begin{figure*}[t]
\centering
\includegraphics[width=\textwidth]{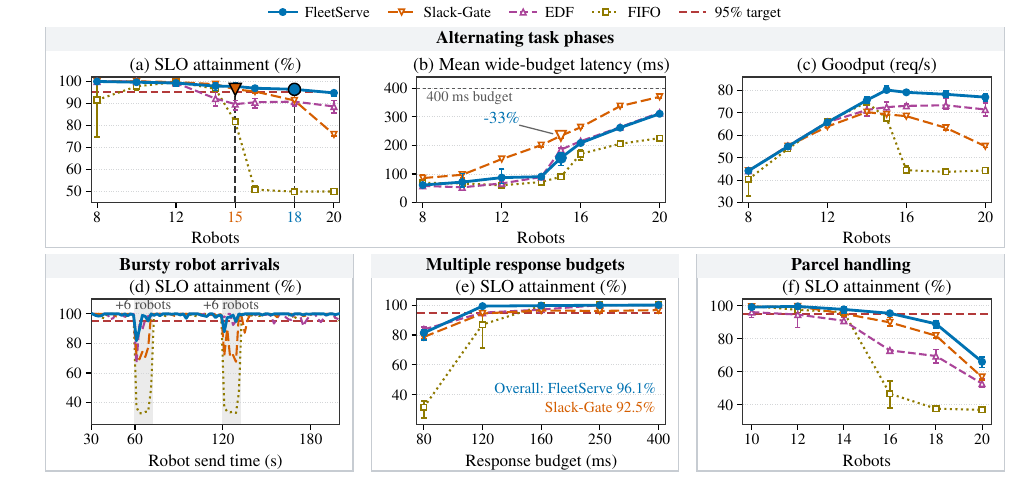}
\caption{Four workloads with matched resources: (a--c) show three metrics for alternating phases;
(d--f) show attainment for the other workloads. Curves show means with min--max whiskers.
Large markers denote ERC in (a) and 33\% lower mean latency for \sys{} versus Slack-Gate
at 15 robots in (b). Shading in (d) marks added robots' measured active periods.}
\label{fig:slo-final}
\end{figure*}

\subsection{Serving-System Comparison}
\label{sec:eval:main}
\label{sec:eval:models}

\sys{} achieves ERCs of 18, 15, and 24 robots across the three models, exceeding
all compared systems (Table~\ref{tab:serving-capacity}). Compared with Robion's
11, 10, and 19 robots, these ERC values are 63.6\%, 50\%, and 26.3\% higher.

For each model, all five systems are evaluated at the same robot counts.
At low load, throughput follows the aggregate request rate; after saturation,
responses delay subsequent sends and latency rises sharply
(Figure~\ref{fig:serving-capacity}). Appendix~\ref{app:serving:capacity}
details ERC validation.

These complete-system comparisons combine multiple mechanisms;
Section~\ref{sec:eval:profile} separates their contributions within a common runtime.

\subsection{Serving Heterogeneous SLOs}
\label{sec:eval:slo}

\paragraph{Alternating task phases.}
In our block-stacking tasks, alignment and contact are more sensitive to delayed
feedback than transport. Motivated by this observation and the recorded phase
patterns (Appendix~\ref{app:slo:phase-traces}), we alternate each robot's response
budget between 90 and 400\ms{} every 30 completed requests, with staggered robot
phases and the closed-loop trigger in
Equation~\ref{eq:request-trigger}. We report SLO attainment, on-time goodput,
and mean wide-budget response latency.

\sys{} achieves an ERC of 18 robots under the 95\% SLO criterion,
compared with 15 for Slack-Gate, 14 for FIFO, and 12 for EDF
(Figure~\ref{fig:slo-final}). EDF degrades gracefully under overload: it keeps
88.6--90.8\% attainment from 16 to 20 robots, where FIFO falls to about 50\%,
and exceeds Slack-Gate at 20 robots (88.6 versus 75.9\%).
Appendix~\ref{app:slo:final} gives detailed results.
At 15 robots, both \sys{} and Slack-Gate satisfy the 95\% overall SLO target.
\sys{} reduces mean wide-budget latency by 33\% and increases goodput from
69.3 to 80.2 responses/s. The closed-loop relation in
Equation~\ref{eq:closed-loop-rates} explains how shorter request cycles can
increase goodput alongside improved attainment.

\paragraph{Bursty robot arrivals.}
To test transient overload, we start with 12 robots and add six robots with
90\ms{} budgets at 60\,s and again at 120\,s, each issuing 55 requests.
Figure~\ref{fig:slo-final}(d) shows the resulting attainment over time.
Within the designated 60--80\,s and 120--140\,s windows, \sys{} attains
96.6\%, versus 95.9\% for EDF and 85.1\% for Slack-Gate.

\paragraph{Multiple response budgets.}
To broaden the budget distribution, 15 robots cycle through 80, 120, 160,
250, and 400\ms{} budgets every 30 responses, with staggered phases.
\sys{} attains 96.1\% overall versus 95.1\% for EDF and 92.5\% for Slack-Gate,
with the per-budget breakdown in Figure~\ref{fig:slo-final}(e).
Appendix~\ref{app:slo:final-sweep} reports per-round results and aggregation details.

\paragraph{Parcel handling.}
The parcel-handling workload derives response-indexed budget sequences from
gripper events in 23 episodes recorded in production. Sequences advance with successful
responses, yielding 61--63\% tight requests and intervals in which all robots
carry tight budgets (Appendix~\ref{app:slo:sorting}). Under the 95\% SLO criterion, \sys{} achieves
an ERC of 14 robots, versus 12 for Slack-Gate and 10 for FIFO. EDF has no confirmed ERC
within the tested range of 10--20 robots
(Figure~\ref{fig:slo-final}(f)). From 16 to 20 robots, observed SLO attainment gains for \sys{}
over Slack-Gate span 3.8--11.4 percentage points, while wide-budget responses
return 7--39\ms{} sooner.

\subsection{Execution Engine Ablations}
\label{sec:eval:profile}
\label{sec:eval:ablation}

\begin{figure*}[t]
\centering
\includegraphics[width=\textwidth]{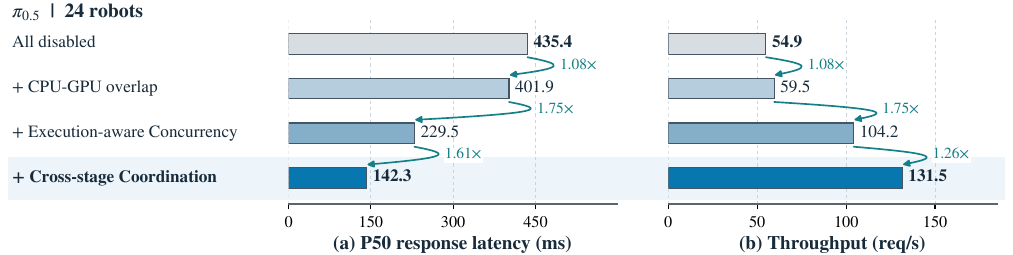}
\caption{Cumulative ablation on $\pi_{0.5}$ at 24 robots.
Bars show per-window means, rows retain prior mechanisms, and arrows show incremental
gains. Execution-aware concurrency enables intra-stage concurrency. Cross-stage
coordination combines Context--Action overlap and independent progression.
Full engine: 2.40$\times$ throughput and 3.06$\times$ P50 speedup over All-Off.}
\label{fig:serving:cumulative}
\end{figure*}

CPU--GPU overlap in Figure~\ref{fig:serving:cumulative} permits host work
during GPU execution, but retains serial GPU execution. Execution-aware concurrency
enables the selected intra-stage multi-stream mode, raising throughput from
59.52 to 104.23\,req/s (1.75$\times$). Cross-stage coordination then combines
Context--Action overlap with independent progression, reaching 131.53\,req/s
(1.26$\times$) and reducing P50 from 229.45 to 142.35\ms{} (1.61$\times$ speedup).
These increments depend on the enable order.

Closed-loop triggering limits throughput headroom. The 180\ms{} minimum send
interval caps 24 robots at $24/0.18\approx133.33$\,req/s, only 27.9\% above the
penultimate configuration. The full engine reaches 131.53\,req/s.
Responses arriving before the next eligible trigger cannot advance the next
request (Equation~\ref{eq:request-trigger}), so latency reductions need not yield
proportional QPS gains. Appendix~\ref{app:serving:mechanisms} details the controls and protocol.

\subsection{Scheduler Analysis}
\label{sec:eval:scheduler}
\label{sec:eval:slo:ablation}

\begin{figure}[t]
\centering
\includegraphics[width=\columnwidth]{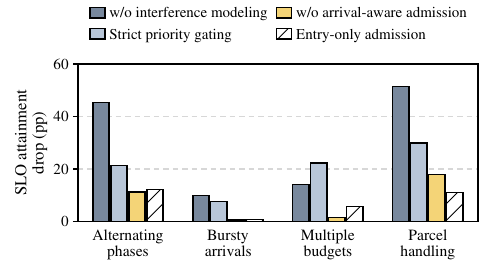}
\caption{Scheduler ablations. Mean SLO attainment drop relative to full
\sys{} for each scheduler variant. Appendix~\ref{app:slo:gpu-ablation}
gives workload settings.}
\label{fig:scheduler-analysis}
\end{figure}

\paragraph{Policy controls and component ablations.}
Figure~\ref{fig:scheduler-analysis} compares four controls with full \sys{}.
Removing interference modeling replaces concurrent-stage predictions with
isolated durations, lowering mean attainment by 45.5 and 51.4 percentage
points for alternating phases and parcel handling at 18 robots.
Strict priority gating blocks ordinary work whenever a protected request
exists, reducing attainment by 7.5--30.5 points across eight settings.
Arrival-aware and stage-wise admission primarily protect tight-budget requests:
disabling arrival-aware checks or checking only before ViT weakens this
protection as load approaches saturation. The losses grow with the tight-budget
share and concentrate within bursts; strict gating instead starves ordinary
work. Appendix~\ref{app:slo:gpu-ablation} reports workload-level results and the
latency and goodput effects of arrival-aware admission.

\paragraph{Running-stage prediction.}
Causal replay of recorded stage transitions at 8/18/20 robots uses only the
concurrency history available at each sample. Mean absolute errors for the
remaining execution time of running stages are 2.0/2.6/2.8\ms{}.
Appendix~\ref{app:slo:online-prediction} details the measurement protocol and
remaining-time error distributions.

\paragraph{Scheduling overhead.}
In chronological CPU replay, a policy update followed by an admission check
takes 73/77\,$\mu$s at P99 with 15/18 robots, below 0.1\% of the 90\ms{}
response budget. This measures CPU policy time, not added response latency;
GPU execution proceeds asynchronously. Appendix~\ref{app:slo:gpu-ablation}
gives the replay protocol.

\section{Related Work}
\label{sec:related}

\paragraph{Robot and VLA serving.}
Prior robot-serving systems use execution context to guide scheduling: Kairos
adapts execution horizons, Armory uses action-queue state, and ROSA coordinates
placement, batching, and request rates \citep{kairos,armory,rosa}.
Robion overlaps encoding and action diffusion with fixed stage batches and
lockstep progression, and selects model queues by their head requests' SLOs
\citep{robion}. \sys{} also exploits stage overlap, but advances requests
independently and separates request priority from interference-aware stage admission.

\paragraph{Stage-aware inference.}
Iteration-level scheduling and chunked prefill improve LLM serving
\citep{orca,sarathi}, while DistServe, Splitwise, and semi-PD separate prefill
and decoding resources \citep{distserve,splitwise,semipd}.
In VLA serving, encoding and denoising use different model components, and
closed-loop control couples response time to subsequent arrivals. Optimized
inference runtimes and model-level work reduction
\citep{flashrt,flashrtpai,embodiedcpp,flashvla} complement our service;
RTC maintains action-chunk continuity during ongoing motion \citep{rtc}.
\sys{} coordinates shared execution using per-request response-time budgets.

\paragraph{SLO-aware GPU sharing.}
Clockwork uses predictable execution to meet request latency targets
\citep{clockwork}. SLOs-Serve allocates work to meet application- and stage-specific
SLOs \citep{slosserve,sola}, while Orion limits interfering kernels and REEF supports
controlled kernel preemption \citep{orion,reef}.
Our admission tests adapt slack stealing and non-preemptive blocking analysis
from real-time scheduling \citep{slackstealing,rta,nonpreemptive}. They use request
progress, calibrated concurrent interference, and arrival history to protect
current and anticipated deadlines at VLA stage boundaries. Issued stages remain
non-preemptive, and feasibility is predicted rather than established by
worst-case timing bounds.

\section{Conclusion}
\label{sec:conclusion}
\label{sec:discussion}

\sys{} is a production-deployed VLA serving system that coordinates shared GPU
computation with the changing feedback needs of robotic execution. Its
asynchronous engine exposes parallelism across requests and stages, while
execution-aware scheduling separates request priority from stage admission
to control interference on current and anticipated urgent requests. Across
three VLA models, \sys{} achieves 1.26--3.75$\times$ the effective robot capacity
of the compared systems under a common response-latency criterion. Under
heterogeneous SLOs, it also improves attainment and on-time goodput. These
results demonstrate how coordinating parallelism and interference around
execution-dependent feedback requirements enables more efficient serving
of robot fleets.

\label{body-end}
\section*{Contributors}
\label{sec:contributors}

Fleetserve is a collaborative effort of the X Square Robot team. The full contributor list is given below; $^{\ast}$ denotes core contributors, $^{\dagger}$ denotes the project lead, and $^{\ddagger}$ denotes the corresponding author.

\vspace{0.5em}
\noindent
Yue Zhang$^{\ast}$, Zicong Ye$^{\ast}$, Lufang Chen$^{\ast}$$^{\dagger}$,
Dance Yang, Mingyang Li, Zhenwei Duan,
Michael Zhao, Ziye Deng, Starrick Liu,
Ruizhi Chengze, YongHua Chen,
SongAo Han, Yu Sun, Ran Yu,
Xiaofan Li, Ruyi Gan, Hao Wang$^{\ddagger}$

\clearpage
\bibliography{references}
\bibliographystyle{mlsys2025}

\section*{AI Usage Disclosure}
\label{sec:ai-usage}
The authors led the system and experimental designs. We used OpenAI
ChatGPT/Codex and Anthropic Claude to generate and iteratively revise
implementation code and experimental, analysis, and plotting scripts based
on the authors' specifications. The tools also assisted with literature
review and drafting and revising multiple manuscript sections. The authors
determine the final designs and content and take full responsibility for
the accuracy and integrity of the paper.

\FloatBarrier
\appendix
\section{Response Latency and Physical Task Efficiency}
\label{app:physical-latency}

\begin{figure*}[t]
\centering
\includegraphics[width=\textwidth]{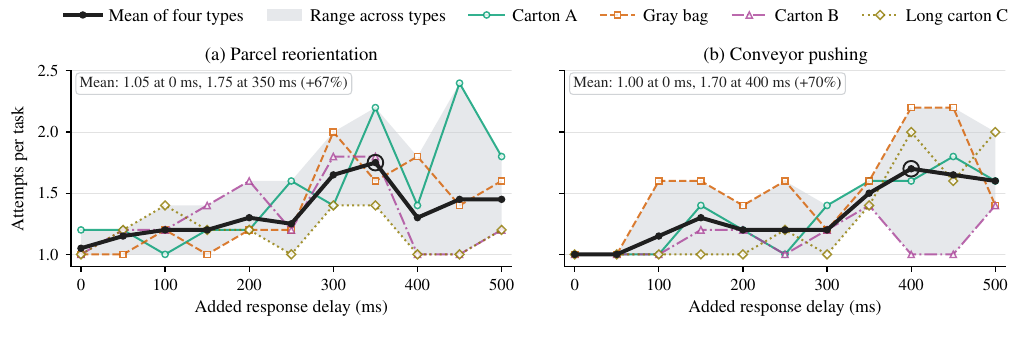}
\caption{Response delay increases physical effort, with different effects across parcel types.
Each point averages five tasks of one type; the black line is the mean of the four types
and the band their range. Zero denotes no added delay. Full counts are in Table~\ref{tab:parcel-attempts}.}
\label{fig:parcel-attempts}
\label{fig:physical-latency}
\end{figure*}

\subsection{Sustained Parcel Handling}
\label{app:physical-output}

\paragraph{Protocol and latency criterion.}
\label{app:physical-budget}
We use one robot and VLA-2B BF16 (\texttt{execute\_horizon}=7), holding
inference settings fixed. Each of six added-delay settings has three five-minute
trials. Delay is inserted between inference completion and response delivery;
baseline response latency is approximately 50\ms{}. Table~\ref{tab:physical-latency}
distinguishes all handled parcels from those reaching the required task endpoint.

\begin{table}[H]
\centering
\small
\caption{Hourly parcel rates under added response delay. Handled and completed
are measured aggregate rates; handled SD is estimated from three trial markers
digitized from the source plot. Figure~\ref{fig:teaser} uses handled rates.}
\label{tab:physical-latency}
\label{tab:physical-digitized}
\setlength{\tabcolsep}{3.5pt}
\begin{tabular}{@{}rrrr@{}}
\toprule
\shortstack{Added delay\\(ms)} & \shortstack{Handled\\parcels/h} &
\shortstack{Completed\\parcels/h} & \shortstack{Handled\\SD} \\
\midrule
0 & 1968 & 1920 & 86.5 \\
25 & 1944 & 1912 & 43.3 \\
50 & 2020 & 1968 & 73.3 \\
100 & 1892 & 1812 & 66.1 \\
150 & 1552 & 1516 & 96.2 \\
200 & 1376 & 1352 & 129.4 \\
\bottomrule
\end{tabular}
\end{table}

Completed output falls 29.6\% at 200\ms{} added delay. Handling throughput drops
6.3\% from 50 to 100\ms{}, then 18.0\% from 100 to 150\ms{}. Selecting 100\ms{}
before this steeper loss, plus the 50\ms{} baseline, motivates the 150\ms{}
response-P50 criterion (Section~\ref{sec:eval:setup}).

\paragraph{Predicted parcel-handling throughput.}
\label{app:physical-projection}
Figure~\ref{fig:teaser} maps serving latency to fleet parcel throughput for 1--16
robots. Let $q(\Delta)$ interpolate the handled rates in
Table~\ref{tab:physical-latency}, and $\widetilde R_s(N)$ interpolate mean per-run
response P50s across robot counts for system $s$ (Figure~\ref{fig:serving-capacity}).
Both interpolations are piecewise linear. Up to 250\ms{} response latency,
\begin{equation}
  \widehat Q_s(N)=N\,q\!\left(\max\{0,\widetilde R_s(N)-50\,\mathrm{ms}\}\right).
\end{equation}
Subtracting 50\ms{} removes baseline latency; faster responses use the
zero-injection rate. FlashRT's measured P50 reaches 505.4\ms{} at 16 robots.
Beyond 250\ms{}, dashed curves extrapolate per-parcel handling time:
\begin{equation}
  q_{\rm ext}(\Delta)=\frac{3600}{t_{200}+k(\Delta-200)},
  \qquad \Delta>200.
\end{equation}
Here $\Delta$ is added delay in milliseconds, $q_{\rm ext}$ is in parcels/h,
and $t_{200}=3600/1376=2.6163$ seconds per parcel. The slope
$k=0.0071353$ seconds per millisecond is fitted to $3600/q(\Delta)$ at
100, 150, and 200\ms{} using only physical calibration data. Adjacent-interval
slope fits give FlashRT 12,137--13,940 parcels/h at 16 robots, so its
comparison with seven FleetServe robots depends on the extrapolation slope.

\paragraph{Uncertainty and scope.}
Using Table~\ref{tab:physical-latency}'s sample SDs $s_j$, we linearly interpolate
variance $v(\Delta)$ within calibration and shade $N[q(\Delta)\pm\sqrt{v(\Delta)}]$,
using $s_0^2$ below baseline. Beyond 250\ms{}, independent sample-SD perturbations
of the calibration means $\bar q_{100}$ and $\bar q_{200}$ propagate through
both the anchor and fitted slope, giving first-order half-width
\begin{equation}
 B(N,\Delta)=N\sqrt{\sum_{j\in\{100,200\}}
       \left(\frac{\partial q_{\rm ext}}{\partial\bar q_j}\right)^2s_j^2}.
\end{equation}
Serving P50s remain fixed. These bands quantify sensitivity to single-robot
variation, not confidence intervals or fleet-level SDs; they exclude
digitization, latency, and model-form uncertainty.

The projection transfers BF16, horizon-7 calibration to MXFP8 Context/BF16 Action
with RTC $d=8$, assuming identical, independent robots with fixed task mix and
control behavior. It treats P50 as constant delay and omits jitter and shared
physical bottlenecks; it is distinct from measured multi-robot task output and ERC.

\subsection{Manipulation Attempts by Parcel Type}
\label{app:physical-attempts}

\paragraph{Protocol.}
We evaluate parcel reorientation (barcode initially down) and conveyor pushing
(barcode initially up) with VLA-2B BF16 and \texttt{execute\_horizon}=7; baseline
model latency is approximately 45\ms{}. Each task tests the four parcel types
in Table~\ref{tab:parcel-attempts}, with five trials per type at each added delay
from 0 to 500\ms{} in 50\ms{} increments. A pushing attempt is one complete push;
injected delay differs from full response latency.

\paragraph{Effect on manipulation effort.}
Attempt counts vary by parcel type and need not increase monotonically with delay
(Figure~\ref{fig:parcel-attempts}). Mean flip attempts rise from 1.05 without added
delay to 1.75 at 350\ms{}; mean push attempts rise from 1.00 to 1.50--1.70 at
350--500\ms{}. Table~\ref{tab:parcel-attempts} gives the complete counts.

\begin{table}[H]
\centering
\small
\caption{Manipulation attempts (flip/push) by parcel type under added response
delay. Type columns give totals over five tasks; Mean gives averages over twenty tasks.}
\label{tab:parcel-attempts}
\label{tab:parcel-flip-attempts}
\label{tab:parcel-push-attempts}
\setlength{\tabcolsep}{2.5pt}
\begin{tabular}{@{}rrrrrr@{}}
\toprule
\shortstack{Added delay\\(ms)} & \shortstack{Carton\\A} &
\shortstack{Gray\\bag} & \shortstack{Carton\\B} &
\shortstack{Long carton\\C} & Mean \\
\midrule
0 & 6/5 & 5/5 & 5/5 & 5/5 & 1.05/1.00 \\
50 & 6/5 & 5/5 & 6/5 & 6/5 & 1.15/1.00 \\
100 & 5/5 & 6/8 & 6/5 & 7/5 & 1.20/1.15 \\
150 & 6/7 & 5/8 & 7/6 & 6/5 & 1.20/1.30 \\
200 & 6/6 & 6/7 & 8/6 & 6/5 & 1.30/1.20 \\
250 & 8/5 & 6/8 & 6/5 & 5/6 & 1.25/1.20 \\
300 & 7/7 & 10/6 & 9/6 & 7/5 & 1.65/1.20 \\
350 & 11/8 & 8/8 & 9/7 & 7/7 & 1.75/1.50 \\
400 & 7/8 & 9/11 & 5/5 & 5/10 & 1.30/1.70 \\
450 & 12/9 & 7/11 & 5/5 & 5/8 & 1.45/1.65 \\
500 & 9/8 & 8/7 & 6/7 & 6/10 & 1.45/1.60 \\
\bottomrule
\end{tabular}
\end{table}


\FloatBarrier

\section{Serving Evaluation: Detailed Results}
\label{app:ablation}
\label{app:serving}

\subsection{Protocol and Configuration}
\label{app:serving:protocol}

\paragraph{Measurement and aggregation.}
Section~\ref{sec:eval:setup} defines the workload and timing boundaries.
Warmup uses the target robot population. The 180\ms{} request period
corresponds to six 30\ms{} model steps in production. Latency statistics
include responses completed after the measurement window for requests sent
within it. We average per-window P50/P95/P99 and QPS; whiskers show the
observed min--max range. Each repeat restarts the client, while the model
service may remain resident.


\paragraph{RTC handoff.}
\label{app:serving:rtc}
After freezing the first $d$ keyframes, RTC \citep{rtc} guides subsequent
keyframes toward the remaining old trajectory with decaying weight and
generates the rest freely. Excessive commitment is \emph{over-freezing};
a response after $t_k+d\Delta$ causes a \emph{handoff violation}, whereas
\emph{starvation} means available actions run out before replacement.
A \emph{trigger miss} occurs when $s_j+\TE$ passes while request $j$
remains in flight (Equation~\ref{eq:request-trigger}).

\paragraph{Input preprocessing.}
VLA-2B and Wall-OSS-0.5 merge image patches in their trained encoders.
\modelB{} instead averages non-overlapping 2$\times$2 patch blocks
after the full ViT, producing 77, 70, and 70 tokens across the three cameras.
Equal visual-token counts do not imply equal total FLOPs.
Its native FlashRT backend uses per-tensor FP8 E4M3 with BF16 inputs and outputs;
Section~\ref{sec:impl} specifies the other precision settings.

The shared instruction retains its original metadata. Wall-X converts,
normalizes, and discretizes the all-zero 14-dimensional state into 20 values,
encoded as 72/74 tokens for VLA-2B/Wall-OSS-0.5.
\modelB{} uses identity normalization and its native state-as-text formatter,
padding to 32 dimensions and discretizing into 256 bins. This synthetic state
is not a calibrated physical-robot state policy. State labels, punctuation,
and special markers belong to the template column of
Table~\ref{tab:serving:inputs}.

\paragraph{Serving interfaces.}
SGLang and vLLM-Omni retain their queues. FlashRT uses its native
single-consumer model session over HTTP with JSON/base64; the other paths
use WebSocket with MessagePack. Transport costs are included in response
latency. \modelB{} enables Context-completion notification.

\subsection{ERC Validation}
\label{app:serving:capacity}

We apply the criterion in Section~\ref{sec:eval:setup} to unrounded
window-level values: every window at the reported ERC must pass, and the
adjacent larger population must fail in at least one window.
For $\pi_{0.5}$, \sys{} passes at 24 robots with P50s of
133.470--136.592\ms{} and mean QPS 132.31; all three windows at 25 robots
fail (163.104--166.842\ms{}). VLA-2B Robion at 12 robots and
$\pi_{0.5}$ Robion at 20 robots each pass in only two of three windows,
despite passing on mean P50.

\subsection{Robion Reproduction Settings}
\label{app:serving:robion}

Table~\ref{tab:serving:robion-settings} specifies our single-GPU
reproduction of Robion~\citep{robion}.

\begin{table*}[!tbp]
\centering\small
\setlength{\tabcolsep}{6pt}
\caption{Robion configuration used in the serving-system comparison.}
\label{tab:serving:robion-settings}
\begin{tabular}{@{}p{0.22\textwidth}p{\dimexpr0.78\textwidth-12pt\relax}@{}}
\toprule
Setting & Configuration \\
\midrule
Scope & Single-model serving; distributed placement and routing are excluded. \\
Execution units & Context and Action each use a sequential CUDA stream with a dedicated CPU submission thread. \\
Round progression & Context work in round $k$ overlaps Action work from round $k-1$; the next round starts after both finish. \\
Result publication & Publish completed Action results after the round barrier (a reproduction choice); CPU output processing may overlap the next GPU round. \\
SM allocation & An offline lookup selects the Context SM quota that minimizes full-round CUDA completion time; Action remains unrestricted. \\
CUDA Graphs & Capture supported execution-unit and Context SM configurations before measurement. \\
CPU preparation & Keep input preparation pipelined with GPU execution. \\
\bottomrule
\end{tabular}
\end{table*}


\subsection{Execution Mechanisms}
\label{app:serving:mechanisms}
\label{app:serving:incremental}

\begin{table}[!b]
\centering\footnotesize
\setlength{\tabcolsep}{2.8pt}
\caption{Incremental ablations on $\pi_{0.5}$ using a multi-stream baseline.
M denotes execution-aware concurrency, C CPU--GPU overlap, R independent progression,
and G GPU-stage overlap. C, R, and G are enabled individually on top of M.
All-Off disables all four mechanisms.
QPS is measured in requests/s. P50, P95, and P99 are client response latencies in ms.}
\label{tab:serving:mechanism-details}
\begin{tabular}{@{}rlrrrr@{}}
\toprule
N & Configuration & QPS & P50 & P95 & P99 \\
\midrule
8 & All-Off & 44.26 & 20.00 & 26.92 & 32.46 \\
 & M & 44.25 & 19.51 & 24.50 & 30.54 \\
 & M+C & 44.26 & 19.87 & 24.82 & 27.34 \\
 & M+R & 44.25 & 20.40 & 31.38 & 33.66 \\
 & M+G & 44.27 & 19.67 & 20.54 & 21.08 \\
 & M+C+R+G & 44.27 & 19.85 & 21.19 & 21.71 \\
\midrule
16 & All-Off & 54.81 & 305.80 & 350.92 & 364.93 \\
 & M & 87.98 & 145.05 & 184.75 & 208.85 \\
 & M+C & 88.44 & 85.00 & 97.11 & 103.96 \\
 & M+R & 87.46 & 152.56 & 193.70 & 224.76 \\
 & M+G & 88.44 & 109.63 & 151.40 & 172.86 \\
 & M+C+R+G & 88.51 & 39.46 & 61.53 & 73.70 \\
\midrule
32 & All-Off & 53.77 & 581.14 & 672.59 & 691.47 \\
 & M & 90.57 & 351.93 & 428.52 & 467.07 \\
 & M+C & 100.79 & 327.11 & 352.05 & 362.57 \\
 & M+R & 99.33 & 311.91 & 461.74 & 560.28 \\
 & M+G & 96.67 & 327.68 & 423.47 & 475.40 \\
 & M+C+R+G & 132.24 & 235.18 & 310.03 & 378.06 \\
\bottomrule
\end{tabular}
\end{table}

\paragraph{Cumulative protocol.}
Figure~\ref{fig:serving:cumulative} follows the sequence in
Section~\ref{sec:eval:profile} at 24 robots with fixed inputs, backend, and
resources. All-Off and all-enabled each use two windows, while intermediate
configurations use one, totaling six displayed windows under
Appendix~\ref{app:serving:protocol}. Values average per-window P50 and QPS.
Execution-aware concurrency enables intra-stage concurrency with the round
barrier retained. Cross-stage coordination combines Context--Action overlap
with independent progression, removing that barrier.
Gains depend on the enabling order. This study does not re-estimate ERC.


\paragraph{Incremental controls.}
Table~\ref{tab:serving:mechanism-details} evaluates six configurations at
8, 16, and 32 robots, each with a separate service start and one window.
All-Off serializes GPU stages, allows one request per stage, and retains
a round barrier. M denotes execution-aware concurrency with the selected
intra-stage streams, concurrency, and batching limits.
C overlaps CPU preparation and output processing with GPU work.
R removes only the round barrier. G overlaps combined
ViT--Prefill (Context) with Action on the GPU.
R and G are the two components of cross-stage coordination.
M+C, M+R, and M+G each add one mechanism to M. All-enabled combines them.
Lockstep states share a round-admission rule, and completed results can
return without waiting for the barrier. Inputs, weights, precision, kernels,
workspaces, KV storage, and captured graphs remain fixed with
completion notification enabled.



\paragraph{Throughput and tail latency.}
At 32 robots, M+R improves QPS and P50 over M while increasing P95/P99.
All-enabled remains above the capacity latency threshold
(Table~\ref{tab:serving:mechanism-details}).


\begin{table}[!htbp]
\centering\footnotesize
\setlength{\tabcolsep}{2.8pt}
\caption{Wall-time shares (\%) at $N=32$ over 61--69\,s.
Ind./Lock.: independent/lockstep; off/on: CPU overlap.
Either/Both: union/intersection of Context/Action kernel intervals.
CPU/GPU: kernel overlap with instrumented CPU preparation/output scopes.}
\label{tab:serving:timeline}
\begin{tabular}{@{}llrrr@{}}
\toprule
Model & State & Either & Both & CPU/GPU \\
\midrule
VLA-2B & Ind., off & 85.49 & 58.36 & 0.00 \\
 & Ind., on & 99.72 & 79.73 & 56.85 \\
 & Lock., on & 90.79 & 25.45 & 11.28 \\
\midrule
Wall-OSS-0.5 & Ind., off & 82.05 & 55.35 & 0.00 \\
 & Ind., on & 98.78 & 87.78 & 26.88 \\
 & Lock., on & 87.55 & 68.99 & 14.49 \\
\midrule
$\pi_{0.5}$ & Ind., off & 72.79 & 19.08 & 0.00 \\
 & Ind., on & 94.73 & 28.80 & 19.60 \\
 & Lock., on & 72.44 & 18.26 & 4.04 \\
\bottomrule
\end{tabular}
\end{table}

\begin{figure}[t]
\centering
\includegraphics[width=\columnwidth]{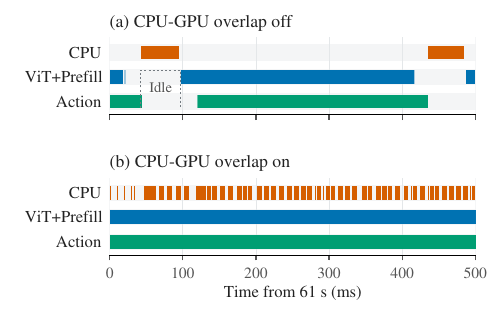}
\caption{CPU--GPU overlap reduces idle gaps in a separate profiler diagnostic. VLA-2B serves
32 robots with overlap (a) disabled and (b) enabled. The figure shows the
61--61.5\,s excerpt; the outlined gap marks an observed GPU idle interval.
Table~\ref{tab:serving:timeline} reports statistics for the full profiling window.}
\label{fig:cpu-gpu-timeline}
\end{figure}

\paragraph{GPU execution timelines.}
Nine separate profiler runs compare three states per model at $N=32$:
independent progression with CPU overlap disabled/enabled, and lockstep
with CPU overlap enabled. Environments and configurations match within
each model; profiler configurations differ from the ablations above. CPU controls toggle preparation/output overlap with GPU work;
progression controls only the round barrier, retaining cross-stage overlap.
Two host-clock markers align the traces; Table~\ref{tab:serving:timeline}
summarizes 61--69\,s and Figure~\ref{fig:cpu-gpu-timeline} shows an excerpt.
With CPU overlap disabled, CPU/GPU intersection is zero at display precision
(below 0.001\% for $\pi_{0.5}$). These shares measure activity, not SM occupancy or MFU;
profiler throughput is excluded from ERC.


\FloatBarrier

\setcounter{dbltopnumber}{3}
\renewcommand{\dbltopfraction}{0.95}
\renewcommand{\dblfloatpagefraction}{0.6}
\section{SLO-Aware Scheduling: Model and Detailed Results}
\label{app:slo}

\subsection{Prediction and Dispatch Model}
\label{app:slo:model}
The central backend schedules non-preemptive units at three boundaries: ViT,
post-vision \phaseP{}, and a complete \phaseA{} denoising trajectory. Its
evaluation uses batch size one. The response-latency SLO $b_j$ is converted
to a server deadline $\ell_j=a_j+b_j-\delta$, where $a_j$ is server arrival
time and $\delta=4\ms{}$ reserves estimated time outside the server. The
request does not carry its robot-side send timestamp.

\paragraph{Progress-aware remaining-time model.}
For stages $s\in\{V,P,A\}$ in pipeline order, define
\begin{equation}
 q_s(\mathbf n)=h_s+\frac{W_s}{\rho_s(\bar{\mathbf n})},\qquad
 \bar n_s=\min(n_s,c_s).
 \label{eq:slo-stage-time}
\end{equation}
$W_s$ is calibrated stage work in isolation-time units, $\rho_s$ is a
dimensionless rate table, and $h_s$ is stage-issue overhead. Every rate lookup
is clipped to the calibrated bounds $\mathbf c$. For a running stage $s$,
the supplied counts include request $j$'s own unit; its rate therefore uses
$\mathbf n^{-j}=[\mathbf n-\mathbf e_s]_+$, where $\mathbf e_s$ is the unit
vector for stage $s$ and the positive part is componentwise. Waiting and
downstream stages use $q_s(\mathbf n)$ without subtraction. In foreseeable
load $\mathbf n_f$, protected ready units already included by the scheduler
remain in this lookup, even for their own requests. The supplied load approximates future concurrency.

Let $\mathbf n_r(u)$ be the scheduler-observed running counts at time $u$,
including request $j$'s own unit, and define
$\mathbf n_r^{-j}(u)=[\mathbf n_r(u)-\mathbf e_s]_+$.
For a stage issued at $\tau_j$, its modeled compute start is
$b_{j,s}=\tau_j+h_s$. Completed work and residual time are
\begin{equation}
\begin{split}
 \widehat w_{j,s}(t)
   &=\int_{b_{j,s}}^{\max(t,b_{j,s})}
        \rho_s(\overline{\mathbf n_r^{-j}(u)})\,du,\\
 \widetilde r_{j,s}(t,\mathbf n)
   &=[b_{j,s}-t]_+
     +\frac{[W_s-\widehat w_{j,s}(t)]_+}
                 {\rho_s(\overline{\mathbf n^{-j}})}.
\end{split}
\label{eq:slo-history-progress}
\end{equation}
Here $[x]_+=\max(0,x)$. The integral is a sum over intervals between
observed stage transitions. Historical rates determine completed work;
the supplied load determines the rate used for the remaining work.
The remaining pipeline time is
\begin{equation}
\begin{split}
 R_j(\mathbf n)&=h_{\mathrm{out}}+r_{j,s}(\mathbf n)
                  +\sum_{k>s}q_k(\mathbf n),\\
 r_{j,s}(\mathbf n)&=
 \begin{cases}
 q_s(\mathbf n), & \text{waiting},\\
 \widetilde r_{j,s}(t,\mathbf n), & \text{running}.
 \end{cases}
\end{split}
\label{eq:slo-remaining}
\end{equation}
The term $h_{\mathrm{out}}$ covers output processing. A completion event
advances the request to its next stage. Future stages use the same projected
load, without an additional prediction of future queueing delay. The counters
describe scheduler in-flight units, including observation delay.

\paragraph{Calibration.}
The rate table comes from a microbenchmark with
RTC $d=8$ and B1 Action trajectories. It measures isolated and
mixed-stage execution at 167 concurrency combinations, repeated three times
per calibration instance. Rates are derived from isolated work divided by measured unit duration.
The frozen table incorporates stage corrections and a common work factor
of 1.02 fitted during simulator calibration. This yields
$(W_V,W_P,W_A)=(4.328931,8.907749,12.116308)\ms{}$.
The model uses $(h_V,h_P,h_A)=(2.1,0.25,1.1)\ms{}$ and
$h_{\mathrm{out}}=1.2\ms{}$; $h_A$ includes preparation and KV copying.
The front-end reserve $f=8.5\ms{}$ is added only for forecast arrivals below.
These parameters were frozen before the reported final-policy GPU comparisons
and are not fitted to their outcomes.


\paragraph{Current-request protection.}
Algorithm~\ref{alg:slo-priority} implements the promotion and protection rules
of Section~\ref{sec:design:priority}, refreshing the effective protected set
$H$ and foreseeable load $\mathbf n_f$ as the promoted set $\mathcal P$ grows.
Ordinary dispatch must preserve every protected request's predicted feasibility;
these predictions are not worst-case timing bounds or SLO guarantees.

\paragraph{Arrival-aware admission.}
Recent arrivals are grouped by observed SLO value. For a value seen $k>0$
times in the rolling window $P=180\ms{}$, forecasts start at
$\max(t,a_{\mathrm{last}}+P/k)$ and repeat every $P/k$, without robot identities
or predefined tight/wide classes. Before issuing ordinary unit $u$,
\textsc{ForecastFeasible} (Section~\ref{sec:design:slo}) checks forecasts during
$u$'s estimated execution, retaining only those with $\ell_e<\ell_u$.
Its projected load $\mathbf n'_e$ adds $u$ to foreseeable load and accumulates
earlier retained ViT forecasts. The front-end reserve is $f=8.5\ms{}$;
physical resources still bound concurrency.

\begin{algorithm}[t]
\caption{Refresh protected requests and foreseeable load}
\label{alg:slo-priority}
\small
\begin{algorithmic}[1]
\REQUIRE State $S$: live requests $\mathcal J$, running units $\mathcal I$,
promoted set $\mathcal P$, and current time $t$
\STATE $\mathcal P\gets\mathcal P\cap\mathcal J$
\STATE $\mathcal L\gets\{j\in\mathcal J:\ell_j-t<R_j(\mathbf0)\}$
\REPEAT
  \STATE $H\gets\mathcal P\setminus\mathcal L$
  \STATE $\mathbf n_f\gets$ \textsc{ForeseeableLoad}$(S,H)$
  \STATE Compute $B_j$ for each $j\in\mathcal J$ from other requests' low-priority units
  \STATE $\mathcal U\gets\{j\in\mathcal J\setminus\mathcal P:\sigma_j(\mathbf n_f)<B_j\}$
  \STATE $\mathcal P\gets\mathcal P\cup\mathcal U$
\UNTIL{$\mathcal U=\emptyset$}
\STATE Save $\mathcal P$ in $S$; \textbf{return} $H,\mathbf n_f$
\end{algorithmic}
\end{algorithm}

\paragraph{Termination of priority refresh.}
With time, live requests $\mathcal J$, and execution state fixed, each nonempty
update only adds members to $\mathcal P$, so the loop terminates within
$|\mathcal J|-|\mathcal P_0|+1$ evaluations including the final empty update,
where $\mathcal P_0$ is the initial promoted set after intersection with
$\mathcal J$, without requiring monotonic calibrated rates.

In Algorithm~\ref{alg:slo-policy}, \textsc{ScanReady} initially orders each
stage's ready work by protected status, deadline, then request ID. After each
launch updates occupancy and refreshes priority, Action retains its remaining
order, while Prefill and ViT rebuild it. Failed checks advance the scan while
leaving candidates queued; both checks apply only to ordinary work.

\subsection{Running-Stage Prediction Accuracy}
\label{app:slo:online-prediction}

\paragraph{Protocol and causal replay.}
We evaluate the remaining execution time of running stages using
Equation~\ref{eq:slo-history-progress}, with one recorded run each at
8, 18, and 20 robots under the alternating 90/400\ms{} workload. We use the
frozen calibration and lookup bounds of Appendix~\ref{app:slo:model}.
At the first policy observation at least 20\ms{} after the preceding sample,
we capture request state and final foreseeable load, without adding predictor
evaluations to the scheduling path. We retain valid running-stage forecasts
for requests sent in $[30,210)$\,s and track each sampled stage through its
observed completion.

Replay integrates scheduler-observed dispatch and retirement transitions strictly
before forecast time $t$ and uses the captured $\mathbf n_f$. Tests verify that
changing future events cannot change a prediction. The artifact retains the
pre-run protocol, instrumentation, analysis scripts, and raw traces.

\paragraph{Stage endpoints and aggregation.}
ViT and Prefill end when completion is observed by the scheduler. Action ends
after its RPC and CPU output processing. The stage-endpoint estimate
$\widehat r_{j,s}(t)$ equals $\widetilde r_{j,s}(t,\mathbf n_f)$ from
Equation~\ref{eq:slo-history-progress}, with the fixed $h_{\mathrm{out}}$ added
for Action. The signed prediction error is
\begin{equation}
 e_{j,s,t}=t^{\mathrm{actual}}_{j,s,\mathrm{end}}-t-\widehat r_{j,s}(t).
\label{eq:slo-prediction-error}
\end{equation}
Positive error denotes underprediction. Primary mean absolute errors (MAEs)
weight each forecast equally. Request-balanced summaries give each represented
request equal total weight within a robot population or stage, divided equally
among its forecasts.

\begin{figure*}[t]
\centering
\includegraphics[width=\textwidth]{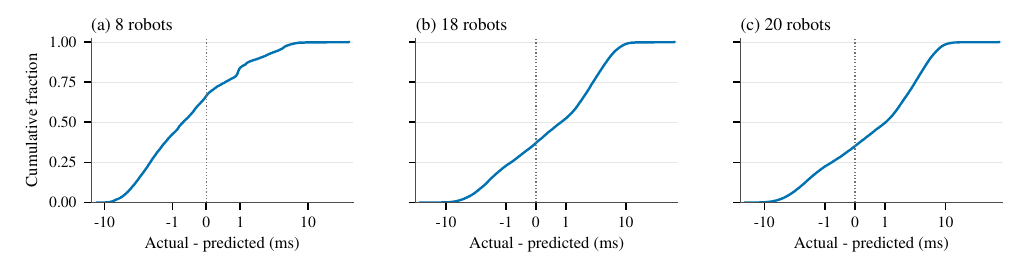}
\caption{Remaining-time prediction errors for running stages.
Positive error is underprediction. Each represented request has equal total
weight within its robot population. Curves include both tails; the horizontal
scale is logarithmic outside $[-1,1]$\ms{}.}
\label{fig:remaining-prediction}
\end{figure*}

\paragraph{Results.}
Across all running-stage samples, current-stage MAE is 2.04/2.62/2.79\ms{} at
8/18/20 robots. With equal request weights within each stage, MAE is
1.91--2.49\ms{} for ViT and Prefill and 2.15--3.05\ms{} for Action.
Figure~\ref{fig:remaining-prediction} shows both error directions.

\subsection{GPU Ablations and Scheduling Cost}
\label{app:slo:gpu-ablation}
\label{app:slo:gpu}
\label{app:slo:ablation}

\paragraph{Recorded task phases.}
\label{app:slo:phase-traces}
Eight recorded $\pi_{0.5}$ block-stacking episodes motivate the alternating
phase workload. Marking 1.0\,s before to 0.5\,s after each gripper switch as
critical covers 15--72\% of each trace. The evaluation abstracts these phase
patterns into controlled 90/400\ms{} budgets and 30-response phases.

\paragraph{Workloads and measurement.}
The GPU controls use the workloads in Section~\ref{sec:eval:slo}, with
batch-size-one execution, a 180\ms{} minimum send interval, and RTC $d=8$.
Appendix~\ref{app:slo:sorting} specifies the parcel-handling trace.
Response latency spans the client send to receipt of the complete response.
We retain requests sent during $[30,210)$\,s through completion and divide
their on-time response count by 180\,s for goodput. Unless stated otherwise,
bursty results cover this full window; burst-only results use 60--80\,s and
120--140\,s. All measured requests complete successfully.

\paragraph{Component controls.}
Five controls change one mechanism or a specified policy combination while
matching the full policy's execution resources.
Removing interference modeling uses isolated-stage durations without changing
actual execution. Strict priority gating replaces both slack stealing and
arrival-aware admission: ordinary work waits whenever a protected request
exists. Removing arrival-aware admission retains current-request protection.
Entry-only admission checks before ViT; Prefill and Action advance in
priority/deadline order when resources permit. Zero-slack promotion sets the
blocking term to zero.

Each of three campaigns rotates policy order across three matched rounds,
with a 75\,s premeasurement smoke run. In Table~\ref{tab:slo:gpu-ablation},
full-policy means pool nine rounds, while each difference uses its own
campaign's three paired rounds. Zero-slack promotion has mixed effects.

\begin{table*}[tp]
\centering\small
\caption{GPU component ablations. Strict gating jointly replaces
slack stealing and arrival-aware admission. FS: SLO attainment of full FleetServe
(mean of the nine rounds in the three campaigns, \%). Other columns: gain of full
FleetServe over each variant (percentage points, mean of three rounds paired with
full FleetServe in the same campaign), with the number of rounds in which full
FleetServe is higher. Bursty robot arrivals use the full measurement window.}
\label{tab:slo:gpu-ablation}
\begin{tabular}{lrrrrrr}
\toprule
Workload & FS & \shortstack{No interference\\model} & \shortstack{Strict\\gating} & \shortstack{No arrival-\\aware adm.} & \shortstack{Entry-only\\admission} & \shortstack{Zero-slack\\promotion} \\
\midrule
Phases, $N=15$ & 96.49 & $+24.29$ (3) & $+7.73$ (3) & $-0.40$ (2) & $+4.87$ (3) & $+0.27$ (1) \\
Phases, $N=16$ & 95.57 & $+45.74$ (3) & $+11.16$ (3) & $+5.02$ (3) & $+15.97$ (3) & $+0.19$ (1) \\
Phases, $N=18$ & 95.96 & $+45.47$ (3) & $+21.34$ (3) & $+11.31$ (3) & $+12.13$ (3) & $+1.04$ (3) \\
Phases, $N=20$ & 94.23 & $+45.68$ (3) & $+30.49$ (3) & $+13.80$ (3) & $+11.07$ (3) & $-0.29$ (1) \\
Bursty arrivals & 98.28 & $+9.92$ (3) & $+7.54$ (3) & $+0.58$ (2) & $+0.71$ (2) & $+2.26$ (3) \\
Multiple budgets & 96.01 & $+14.24$ (3) & $+22.29$ (3) & $+1.47$ (2) & $+5.64$ (3) & $-0.25$ (2) \\
Parcel, $N=16$ & 94.78 & $+56.62$ (3) & $+18.55$ (3) & $+16.97$ (3) & $+13.77$ (3) & $-0.77$ (1) \\
Parcel, $N=18$ & 88.77 & $+51.44$ (3) & $+29.90$ (3) & $+18.05$ (3) & $+11.12$ (3) & $+2.01$ (2) \\
\bottomrule
\end{tabular}
\end{table*}

\paragraph{Arrival-aware admission.}
A separate experiment isolates arrival-aware admission in six matched rounds
with rotated policy order and identical model instances, stage computations,
and execution resources. These rounds are not pooled with the component
campaigns. Table~\ref{tab:slo:t69summary} retains both attainment and the latency
and goodput effects: protecting tight-budget requests can increase wider-budget
latency and need not increase goodput.

\begin{table*}[t]
\centering\small
\setlength{\tabcolsep}{8pt}
\caption{Arrival-aware GPU ablation, six-round means. Each cell gives
U / T: FleetServe without / with arrival-aware admission under matched
execution resources. Tight budgets are 90\ms{} for alternating phases and
bursty arrivals, and 80\ms{} for multiple budgets. Wide-budget latency is for
the 400\ms{} class. Bursty results use the full measurement window.}
\label{tab:slo:t69summary}
\begin{tabular}{lrrrr}
\toprule
Workload & \shortstack{Overall attainment\\(\%)} &
\shortstack{Tight attainment\\(\%)} &
\shortstack{Wide mean latency\\(ms)} & \shortstack{Goodput\\(responses/s)} \\
\midrule
Phases, $N=15$ & 97.3 / 97.8 & 94.5 / 95.6 & 114 / 148 & 80.4 / 80.4 \\
Phases, $N=16$ & 89.2 / 96.6 & 78.3 / 93.2 & 200 / 208 & 74.0 / 79.0 \\
Phases, $N=18$ & 84.3 / 96.5 & 68.7 / 93.0 & 260 / 262 & 68.7 / 78.4 \\
Bursty arrivals & 97.7 / 97.8 & 95.7 / 96.0 & 94 / 95 & 66.6 / 66.5 \\
Multiple budgets & 94.6 / 96.1 & 75.0 / 81.9 & 139 / 236 & 77.7 / 74.7 \\
\bottomrule
\end{tabular}
\end{table*}

\paragraph{CPU decision cost.}
On a Xeon Gold 6530, we replay queued decision states from 40--45\,s of the
simulation four times in chronological order, with no serving process running.
For 8/15/18 robots, FleetServe's policy update and admission check
(including arrival observation) take
41/63, 52/73, and 55/77\,$\mu$s (median/P99), respectively; these are CPU
policy times, not added response latency.

\subsection{Scheduling Policy Comparison}
\label{app:slo:final}
\label{app:slo:final-sweep}
\label{app:slo:pending}

\paragraph{Baseline configuration.}
\label{app:slo:threshold}
All policies share allocated execution resources and per-request stage computation.
Simulation selects EDF's smallest per-stage concurrency limits preserving
the best SLO attainment under alternating task phases at 15 robots; these
limits are fixed for GPU evaluation. Slack-Gate's 40\ms{} threshold is selected
from 20, 40, and 80\ms{} jointly with ViT concurrency, using the same workload
and bursty arrivals. Among candidates within 1 percentage point of the best
attainment, selection favors lower mean latency.

\paragraph{Comparison and aggregation.}
Each policy uses three rounds with a fixed configuration. FIFO and Slack-Gate
rotate within matched rounds; EDF and FleetServe are measured separately.
Comparisons use equally weighted round means, with reused measurements counted
once. Reference-load checks under unchanged configurations show mean-attainment
variation of $-1.02$ to $+1.00$ percentage points. Alternating task phases use
1,200 requests per robot; measurement boundaries follow
Appendix~\ref{app:slo:gpu}. Table~\ref{tab:slo:t82sweep} reports per-round
attainment and differences in three-round mean goodput.

\paragraph{ERC.}
ERC is the largest tested population with at least 95\% attainment in each
of three rounds: FleetServe passes at 18 robots but not at 20; 19 was not
measured. This criterion does not imply that every smaller population passes.

\paragraph{Temporal and budget aggregation.}
Figure~\ref{fig:slo-final}(d) averages attainment in 2\,s request-send-time bins
across rounds. Section~\ref{sec:eval:slo} reports attainment over the 60--80\,s
and 120--140\,s burst windows; Table~\ref{tab:slo:t82sweep} uses the full
$[30,210)$\,s send window. For multiple response budgets
(Figure~\ref{fig:slo-final}(e)), overall attainment aggregates all requests
within each round before averaging rounds, rather than weighting budget
classes equally.

\begin{table*}[p]
\centering
\caption{Scheduling policy comparison under matched execution resources.
SLO attainment (\%) is reported as R1 / R2 / R3. The $N$ rows use alternating
task phases; bursty arrivals use the full $[30,210)$\,s send window.
$\Delta$goodput is the difference between three-round means
(FleetServe $-$ Slack-Gate; on-time responses/s).}
\label{tab:slo:t82sweep}
\footnotesize
\setlength{\tabcolsep}{3pt}
\begin{tabular}{lrrrrr}
\toprule
Workload & FIFO & EDF & Slack-Gate & FleetServe & \shortstack{$\Delta$ goodput\\(responses/s)} \\
\midrule
$N=8$ & 74.86 / 99.96 / 99.81 & 99.86 / 99.92 / 99.99 & 99.97 / 100.00 / 99.90 & 99.94 / 99.91 / 99.94 & $-0.03$ \\
$N=10$ & 98.94 / 97.98 / 96.17 & 99.94 / 99.92 / 99.98 & 99.98 / 99.97 / 99.96 & 99.90 / 98.95 / 99.93 & $-0.14$ \\
$N=12$ & 99.77 / 99.75 / 99.58 & 99.66 / 98.45 / 99.70 & 99.92 / 99.86 / 99.27 & 99.89 / 97.99 / 99.79 & $+1.86$ \\
$N=14$ & 96.30 / 96.01 / 97.49 & 91.05 / 88.72 / 96.71 & 98.86 / 99.26 / 97.93 & 95.82 / 98.22 / 99.48 & $+5.34$ \\
$N=15$ & 81.19 / 81.36 / 82.35 & 87.00 / 90.61 / 91.35 & 95.88 / 97.28 / 95.88 & 96.85 / 97.14 / 99.19 & $+10.84$ \\
$N=16$ & 50.14 / 49.98 / 52.38 & 92.20 / 91.23 / 88.00 & 95.69 / 95.68 / 94.77 & 96.12 / 96.89 / 97.59 & $+10.62$ \\
$N=18$ & 49.97 / 49.99 / 50.00 & 88.92 / 89.42 / 94.05 & 89.99 / 90.33 / 93.63 & 96.12 / 95.56 / 97.43 & $+14.94$ \\
$N=20$ & 50.01 / 49.99 / 50.02 & 85.71 / 88.87 / 91.22 & 75.28 / 75.43 / 76.99 & 95.93 / 93.43 / 94.90 & $+21.90$ \\
\shortstack[l]{Bursty robot\\arrivals} & 88.89 / 89.00 / 88.37 & 98.81 / 97.99 / 97.22 & 95.61 / 95.61 / 97.50 & 98.66 / 98.79 / 99.27 & $+4.08$ \\
\shortstack[l]{Multiple response\\budgets, $N=15$} & 85.49 / 86.31 / 79.02 & 94.13 / 94.80 / 96.44 & 92.16 / 92.47 / 92.79 & 96.61 / 95.01 / 96.58 & $+12.28$ \\
\bottomrule
\end{tabular}
\par\bigskip
\normalsize
\centering
\caption{Parcel handling: overall SLO attainment (\%) as R1 / R2 / R3
under matched execution resources.}
\label{tab:slo:sorting}
\footnotesize
\setlength{\tabcolsep}{3.5pt}
\begin{tabular}{lrrrr}
\toprule
Robots & FIFO & EDF & Slack-Gate & FleetServe \\
\midrule
10 & 99.79 / 99.84 / 99.68 & 99.79 / 92.78 / 95.77 & 99.88 / 99.91 / 98.35 & 98.07 / 99.76 / 99.74 \\
12 & 98.79 / 99.74 / 94.03 & 97.48 / 99.60 / 86.81 & 99.26 / 99.70 / 99.83 & 99.81 / 99.26 / 99.66 \\
14 & 97.46 / 96.57 / 94.39 & 90.26 / 89.49 / 93.02 & 95.87 / 95.80 / 93.89 & 95.49 / 99.10 / 98.67 \\
16 & 54.46 / 38.09 / 47.65 & 72.45 / 73.19 / 73.24 & 89.25 / 87.64 / 92.76 & 95.21 / 94.50 / 96.51 \\
18 & 37.65 / 37.82 / 37.65 & 65.32 / 69.96 / 73.45 & 81.46 / 80.64 / 83.59 & 86.68 / 89.43 / 90.97 \\
20 & 36.94 / 37.12 / 36.84 & 50.65 / 52.13 / 55.79 & 56.90 / 55.18 / 58.40 & 62.66 / 66.61 / 68.99 \\
\bottomrule
\end{tabular}
\end{table*}



\subsection{Parcel handling}
\label{app:slo:sorting}
Client budgets follow recorded contact timing during parcel picking, flipping,
and passing. We use 23 validation episodes, excluding
one collection site with constant gripper readings that obscure contact timing.
A contact interval spans 1.0\,s before to 0.5\,s after any gripper switch, and overlapping
intervals merge; contact covers 36--78\% of each episode (64\% on average).
Robot $i$ replays episode $i \bmod 23$ from a seeded start offset and loops it.
Contact intervals carry the 90\ms{} budget and the remaining time the 400\ms{}
budget. Interval durations convert to response counts at the 180\ms{} period,
because phases advance by successful responses as in alternating task phases. Each robot sends 1,200
requests with the same staggered starts, and each of up to 20 robots replays a
different episode. Tight requests make up 61--63\% of each scenario, and the
fraction of robots in contact varies between 20\% and 100\%.

FIFO and Slack-Gate are measured at 10--20 robots in three rounds; EDF comes
from its separate campaign and FleetServe from a separate run, all with the
protocol of Appendix~\ref{app:slo:final}. Repeated runs of the same configuration differ by up to
2.9 percentage points in three-round mean at 16 and 18 robots, so comparisons
across runs carry uncertainty of this size. All 941,655 requests sent in the
measurement windows succeed. Table~\ref{tab:slo:sorting} reports every round.

At 16 robots, tight-budget attainment is
56.6\% for EDF, 92.6\% for FleetServe, and 83.8\% for Slack-Gate.
From 14 robots upward, FleetServe returns wide-budget responses 7--56\ms{}
sooner than Slack-Gate and increases goodput by 4.1--9.1 requests/s.

\end{document}